%% file: main.tex
\documentclass[runningheads]{llncs}

\PassOptionsToPackage{table}{xcolor}
\PassOptionsToPackage{xcdraw}{xcolor}
 
\usepackage{eccv}

\usepackage{eccvabbrv}

\usepackage{graphicx}
\usepackage{booktabs}

\usepackage[accsupp]{axessibility}  

\usepackage{orcidlink}

\usepackage{hyperref}

\usepackage{orcidlink}
\usepackage{kotex}
\usepackage{xcolor}         
\usepackage{multirow}
\usepackage{makecell}
\usepackage{graphicx}
\usepackage{tabularx}      
\usepackage{wrapfig}
\usepackage{xspace}
\usepackage[capitalize]{cleveref}
\usepackage{enumitem}

\newcommand{\paragrapht}[1]{\noindent\textbf{#1}}

\newcommand{\NA}{\textcolor{lightgray}{\footnotesize N/A}}

\newcommand{\tworow}[2]{\begin{tabular}[c]
{@{}c@{}}#1\vspace{-2pt}\\#2\end{tabular}}
\newcommand{\tbf}[1]{\textbf{#1}}
\newcommand{\tul}[1]{\underline{#1}}

\creflabelformat{equation}{#2#1#3}

\input{math_command}

\makeatletter
\renewcommand*{\@fnsymbol}[1]{\ensuremath{\ifcase#1\or \star \or \dagger \or \ddagger \or \mathsection \or \mathparagraph \or \|\or \star\star \or \dagger\dagger \or \ddagger\ddagger \else\@ctrerr\fi}}
\makeatother

\begin{document}

\title{Relaxed Rigidity with Ray-based Grouping for Dynamic Gaussian Splatting}


\author{
Junoh Lee\inst{1}\thanks{Work done during an internship at NAVER AI Lab.}    \orcidlink{0009-0006-0860-8469}  \and
Junmyeong Lee\inst{2}  \orcidlink{0009-0004-1375-2364}  \and
Yeon-Ji Song\inst{3}   \orcidlink{0009-0009-7412-056X}  \and
Inhwan Bae\inst{4}     \orcidlink{0000-0003-1884-2268}  \and
Jisu Shin\inst{1}      \orcidlink{0009-0007-7640-1512}  \and
Hae-Gon Jeon\inst{2}\thanks{Corresponding authors}  \orcidlink{0000-0003-1105-1666}  \and
Jin-Hwa Kim\inst{3,5}\textsuperscript{$\dagger$}  \orcidlink{0000-0002-0423-0415}}

\authorrunning{L. Junoh et al.}

\institute{
\mbox{}\inst{1}GIST \quad \inst{2}Yonsei Univ. \quad \inst{3}SNU \quad \inst{4}DGIST \quad \inst{5}NAVER AI Lab \\
\mbox{}\email{
\mbox{}\{juno, jsshin98\}@gm.gist.ac.kr, 
\mbox{}\{june65, earboll\}@yonsei.ac.kr, \\
\mbox{}yjsong@snu.ac.kr, 
\mbox{}inhwanbae@dgist.ac.kr, \\
\mbox{}j1nhwa.kim@navercorp.com}}
\maketitle

\input{sections/0_abstract}    
\input{sections/1_intro}
\vspace{-2mm}
\input{sections/2_related}
\vspace{-2mm}
\input{sections/3_method}

\vspace{-2mm}
\input{sections/4_experiment}
\vspace{-2mm}
\input{sections/5_conclusion}

%
%


\clearpage
\bibliographystyle{splncs04}
\bibliography{main}

\input{sections/X_suppl}
\end{document}

%% file: math_command.tex
\usepackage{amsmath,amsfonts,bm}

\def\eqref#1{equation~\ref{#1}}

\def\1{\bm{1}}

\def\vd{{\bm{d}}}

\def\vo{{\bm{o}}}

\DeclareMathAlphabet{\mathsfit}{\encodingdefault}{\sfdefault}{m}{sl}
\SetMathAlphabet{\mathsfit}{bold}{\encodingdefault}{\sfdefault}{bx}{n}



%% file: sections/0_abstract.tex
\begin{abstract}
The reconstruction of dynamic 3D scenes using 3D Gaussian Splatting has shown significant promise. A key challenge, however, remains in modeling realistic motion, as most methods fail to align the motion of Gaussians with real-world physical dynamics. This misalignment is particularly problematic for monocular video datasets, where failing to maintain coherent motion undermines local geometric structure, ultimately leading to degraded reconstruction quality. Consequently, many state-of-the-art approaches rely heavily on external priors, such as optical flow or 2D tracks, to enforce temporal coherence.
In this work, we propose a novel method to explicitly preserve the local geometric structure of Gaussians across time in 4D scenes. Our core idea is to introduce a view-space ray grouping strategy that clusters Gaussians intersected by the same ray, considering only those whose $\alpha$-blending weights exceed a threshold. We then apply constraints to these groups to maintain a consistent spatial distribution, effectively preserving their local geometry. This approach enforces a more physically plausible motion model by ensuring that local geometry remains stable over time, eliminating the reliance on external guidance. We demonstrate the efficacy of our method by integrating it into two distinct baseline models. Extensive experiments on challenging monocular datasets show that our approach significantly outperforms existing methods, achieving superior temporal consistency and reconstruction quality.
\keywords{Dynamic 3D Gaussian Splatting \and Point Grouping \and Motion Regularization}
\end{abstract}

%% file: sections/1_intro.tex
\section{Introduction}
\label{sec:intro}
\vspace{-2mm}

Reconstructing dynamic scenes from videos is essential for modeling spatio-temporal structure in real-world scenes. The emergence of 3D Gaussian Splatting (3DGS)~\cite{kerbl20233d} has significantly advanced this task with its explicit representation and efficient training and rendering. Various dynamic extensions of 3DGS~\cite{wu20234dgs,yang2023real,li2023spacetime} have demonstrated impressive reconstruction, even under monocular camera settings. For instance, these approaches enable continuous spatio-temporal scene modeling by parameterizing time-varying Gaussian primitives.

However, accurately reflecting the physical motion of the real 4D world remains challenging, as the estimated motion is often under- or over-constrained and not directly governed by physical plausibility. To mitigate this issue, prior works~\cite{chen2025freegaussian,zhu2024motiongs} employ motion priors derived from optical flow and depth or impose rigidity assumptions~\cite{huang2023sc,lin2023gaussian,kratimenos2024dynmf}. Nevertheless, these approaches exhibit fundamental limitations. External prior models~\cite{Xie2025GSLK,wang2025som} are not inherently designed to guide the point dynamics of Gaussian primitives. Specifically, these methods project Gaussian motion onto the image plane and rely on optical or 3D flow as proxy guidance. Because such supervision is defined in 2D screen space rather than in the underlying 3D geometry, it provides only indirect and potentially inconsistent motion cues. As a result, errors and ambiguities in the proxy signals are propagated to the optimization process, leading to inaccurate learning of Gaussian motion~\cite{zhu2024motiongs, Xie2025GSLK}.
In contrast, rigid-motion-based models use K-Nearest Neighbors (KNN) to group adjacent points and enforce shared rigid transformations. This strategy overlooks the inherently non-rigid and interactive nature of real-world motion and fails to account for varying spatial scales of Gaussian primitives, which require adaptive grouping regions. Consequently, relying on external priors or strict rigidity assumptions is insufficient to ensure physically consistent 4D representations.

\begin{figure*}[t]
\centering
 \includegraphics[width=0.9\textwidth]{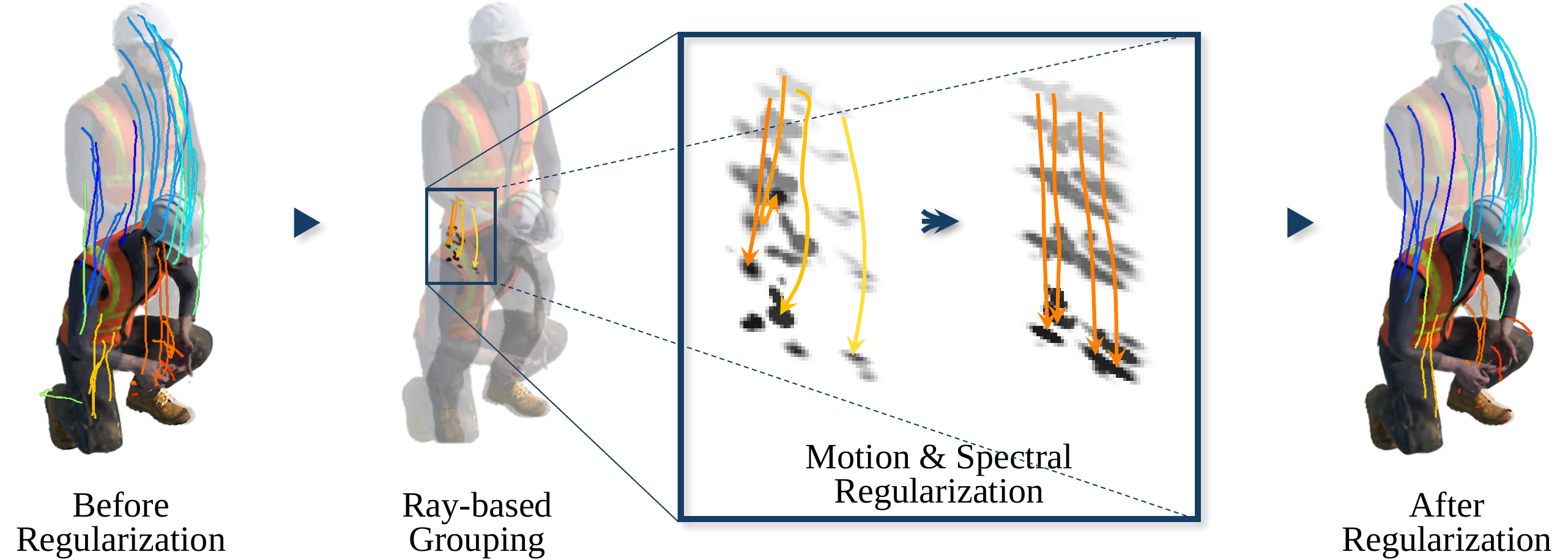}
 \vspace{-2mm}
 \caption{Overview.
Before regularization (first), individual Gaussians exhibit temporally inconsistent motion; each colored trajectory denotes the time-varying movement of a single Gaussian. Our method first performs ray-based grouping to cluster spatially adjacent Gaussians along view rays (second), with a zoomed-in example shown in the third panel. 
Motion and spectral regularization are then applied within each group to enforce coherent dynamics. 
After regularization (fourth), the Gaussian trajectories become temporally aligned and physically consistent.}  
 \label{fig:teaser}
 \vspace{-6mm}
\end{figure*}

To enforce physically plausible constraints, we focus on two key aspects (\cref{fig:teaser}). First, we group spatially adjacent Gaussians that exhibit coherent motion using a ray-based strategy rather than distance-based grouping. For each pixel, we consider only Gaussians intersected by the same view ray whose $\alpha$-blending weights~\cite{kerbl20233d} exceed a threshold, excluding irrelevant primitives along the ray. This corresponds to the sorted and blended Gaussians at the pixel in 3DGS. By selecting only visible, high-contribution Gaussians, we obtain motion-consistent groups that naturally reflect scale, opacity, and position without additional computational overhead during rasterization. Second, we introduce a relaxed rigidity constraint within each group. Rather than enforcing identical 3D displacements, we preserve directional consistency and temporal distribution, allowing non-rigid deformations, while preventing artifacts such as geometric inconsistency and floaters caused by incoherent motions.

To demonstrate the versatility of our approach, we integrate our regularization into four representative 4DGS frameworks: RTD~\cite{wu20234dgs}, Ex4DGS~\cite{lee2024fully}, MoDec-GS~\cite{kwak2025modecgs}, and Grid4D~\cite{xu2024grid4d}. We evaluate on both the synthetic D-NeRF~\cite{pumarola2020dnerf} dataset and real-world benchmarks, HyperNeRF~\cite{park2021hypernerf} and NeRF-DS~\cite{yan2023nerfds}. 
Across all baselines and datasets, our method consistently improves rendering quality, achieving state-of-the-art performance. On the D-NeRF dataset, our approach improves PSNR by an average of $1.19$ dB over the baselines. While the proposed method increases training time by around $2$ to $3$ times, it does not alter the underlying model architectures and therefore introduces no additional cost during rendering.

\vspace{-2mm}
\begin{itemize}[label=\textbullet]
    \item We propose a framework that enforces physically plausible motion without relying on external priors.

    \item We introduce a model-agnostic ray-based grouping strategy and relaxes rigidity constraints, enabling the flexible representation of motion.
    
    \item  We demonstrate that our approach achieves state-of-the-art performance on both synthetic and real-world benchmarks, including D-NeRF, HyperNeRF, and NeRF-DS.
\end{itemize}
\vspace{-2mm}

%% file: sections/2_related.tex
\section{Related Work}
\label{sec:related}
\vspace{-2mm}

\subsection{Monocular Video Novel View Synthesis}
\vspace{-2mm}
Novel view synthesis~\cite{mildenhall2020nerf,lee2024geometry,kerbl20233d} generates unseen views from input images. While extending this task to dynamic scenes relies on multiview videos~\cite{Sabater2017,li2022neural,hyperreel,xu20234k4d,lombardi2019neural,song2023nerfplayer,fridovich2023k,cao2023hexplane,wang2023mixed}, utilizing monocular videos~\cite{park2021nerfies,park2021hypernerf,fang2022TiNeuVox,park2023tidnerf} serves as a highly practical alternative. However, monocular inputs lack cross-view motion cues. Consequently, recent approaches~\cite{zhu2024motiongs,Xie2025GSLK,wang2025som,liu2025modgs,shin2025chroma} employ prior models for optimization. These methods typically leverage optical flow guidance~\cite{teed2020raft,xu2022gmflow}, 2D point tracks~\cite{karaev23cotracker,Xiao2024SpatialTracker}, and depth estimation~\cite{Ranftl2022midas,piccinelli2024unidepth,hu2025DepthCrafter} to guide 4D scene reconstruction. A critical limitation of these approaches is their strong dependency on the accuracy of the underlying priors. When these external models fail, such as in environments with view-dependent artifacts or textureless backgrounds~\cite{zhu2024motiongs,Xie2025GSLK}, the resulting synthesis quality degrades. Therefore, designing physically plausible motion guidance that operates independently of prior models is essential.

\vspace{-2mm}
\subsection{Dynamic Extension of 3DGS}
\vspace{-2mm}
The computational efficiency of 3DGS~\cite{kerbl20233d,hyung2024effective} drives the extension of the method to dynamic scenes. Dynamic 3DGS methodologies are classified into three categories. The first category utilizes deformation fields~\cite{yang2023deformable,lu20243d,sun20243dgstream,wu20234dgs,wan2024spgs,Li2024ST4DGS,bae2024ed3dgs,gao2024hicom,zhu2024motiongs,shaw2024swings,wuswift4d,xu2024grid4d,liang2025himor,kwak2025modecgs} to compute temporal offsets for primitive parameters. The second category employs basis functions. These models adopt Fourier series~\cite{katsumata2024compact}, polynomials~\cite{li2023spacetime,hu2025learnable,wang2025freetimegs} or splines~\cite{lee2024fully,yoon2025splinegs,park2025splinegs}. The third category defines primitives in four-dimensional space~\cite{yang2023real,duan20244drotor,zhang2025mega}. With the exception of native 4D formulations, these methods generate 3D Gaussian primitives to process the primitives through the 3DGS rasterizer. However, regulating the motion of Gaussians is highly ill-posed, necessitating spatial constraints or grouping mechanisms to enforce coherent movement. To achieve this, our approach leverages the rasterization process as a grouping function. This design allows our method to integrate into existing dynamic models without requiring structural modifications.

\vspace{-2mm}
\subsection{Motion Constraints for Dynamic Gaussian Splatting}
\vspace{-2mm}
Imposing motion constraints on Gaussian primitives relies on rigidity assumptions~\cite{luiten2023dynamic, das2023neural, bae2024ed3dgs,wu20254dfly,lee2026space}. Existing methods maintain geometric structure by preserving distances among KNN groups or by enforcing local rigidity through As-Rigid-As-Possible (ARAP) formulations~\cite{sorkine2007arap, huang2023sc, lin2023gaussian, Li2024ST4DGS}. However, the KNN approach~\cite{luiten2023dynamic, huang2023sc, lin2023gaussian, Li2024ST4DGS, bae2024ed3dgs,wu20254dfly} groups Gaussians based on Euclidean distance, which ignores the properties of Gaussian primitives, including scale and opacity. Furthermore, strict rigidity assumptions fail during topological changes~\cite{park2021hypernerf}, where the underlying distances between Gaussians change. Therefore, we propose a novel ray-based grouping strategy and a relaxed rigidity constraint to capture the properties of Gaussian primitives while remaining highly adaptable to complex, non-rigid physical movements.

%% file: sections/3_method.tex
\section{Method}
\vspace{-2mm}

\subsection{Preliminary}
\label{subsec:prelim}
\vspace{-2mm}

\textbf{3DGS}. Our approach builds upon the point-based differentiable rasterization framework of 3DGS~\cite{kerbl20233d}.
3DGS represents a 3D scene with anisotropic Gaussians parameterized by position $\bm{\mu}$, covariance matrix $\bm{\Sigma}$, opacity $\vo$, and color $\bm{c}$.
The covariance $\bm{\Sigma}$ is parameterized via a diagonal scaling matrix $\bm{S}$ (from a scale vector $\bm{s}$) and a rotation matrix $\bm{R}$ (from a quaternion $\bm{q}$):
\vspace{-1mm}
\begin{equation}
\smash{\bm{\Sigma} = \bm{R}\bm{S}\bm{S}^\top\!\bm{R}^\top}.
\end{equation}
To render an image, 3D Gaussians are projected to the image plane using an approximated graphics pipeline~\cite{zwicker2001ewa}.
The 2D covariance $\bm{\Sigma}'$ in camera coordinates is computed as:
\vspace{-1mm}
\begin{equation}
\bm{\Sigma}' = \bm{J}\,\bm{W}\bm{\Sigma}\bm{W}^{\!\top}\!\bm{J}^{\!\top},
\end{equation}
where $\bm{J}$ is the Jacobian of the affine approximation for perspective projection, and $\bm{W}$ is the viewing transformation matrix.
By discarding the third row and column of $\bm{\Sigma}'$, the 3D Gaussian is approximated as a 2D anisotropic Gaussian on the image plane.
The density of the projected Gaussian is:
\vspace{-1mm}
\begin{equation}
\mathcal{G}(\bm{x}) = e^{-\frac{1}{2}(\bm{x}-\bm{\mu})^\top\bm{\Sigma}^{'-1}(\bm{x}-\bm{\mu})}.
\end{equation}
The final pixel color is computed using point-based $\alpha$-blending, analogous to volume rendering.
In 3DGS, $\alpha_i = \vo_i \mathcal{G}(\bm{x})$.
For Gaussians sorted along a ray, the accumulated color $C$ is defined as follows:
\vspace{-1mm}
\begin{equation}
C = \sum_{i=1}^{N}T_{i}(1-e^{-\alpha_i })\bm{c}_i \quad\text{s.t.} \quad T_i=e^{-\sum_{j=1}^{i-1}\alpha_j},
\label{eq:alpha_comp}
\end{equation}
where $N$ is the number of visible Gaussians intersecting the ray, and $i$ is the sorted index by depth.
We define the contribution weight as $w_i = T_i (1 - e^{-\alpha_i})$.

\paragrapht{Deformation-field Dynamic 3DGS Model.}
Dynamic Gaussian Splatting extends 3DGS by introducing time-dependent offsets to the canonical Gaussian parameters. These offsets are typically predicted via a deformation network conditioned on time and the canonical attributes. For a canonical Gaussian $G_c = ( \bm{\mu}_c, \bm{q}_c, \bm{s}_c, \vo_c )$, the deformation field outputs the time-varying offsets such that the per-frame primitives become:
\vspace{-1mm}
\begin{equation}
G_{t} = (\bm{\mu}_c+\Delta\bm{\mu}_t, \bm{q}_c+\Delta\bm{q}_t, \bm{s}_c+\Delta\bm{s}_t, \vo_c+\Delta{\vo}_t),
\label{eq:deform_field}
\end{equation}
where $\Delta(\cdot)$ denotes time-dependent offsets.

\paragrapht{Motion Basis for Dynamic 3DGS Model.}
Basis-trajectory approaches~\cite{li2023spacetime,katsumata2024compact,lee2024fully} parameterize most motions through linear combinations of low-dimensional temporal basis. For example, the position $\bm{\mu}$ can be represented below:
\vspace{-1mm}
\begin{equation}
\bm{\mu}_{i}(t) = \sum_{\ell=1}^{L}\bm{a}_{i,\ell}\,\phi_{\ell}(t),
\label{eq:basis_traj}
\end{equation}
where $\{\phi_{\ell}\}$ are basis functions and $\bm{a}_{i,\ell}$ are learnable coefficients.
In our experiments, we validate our approach on both deformation-field (\cref{eq:deform_field}) and basis-trajectory (\cref{eq:basis_traj}) formulations to demonstrate its applicability across representative dynamic 3DGS architectures.

\vspace{-2mm}
\subsection{Ray-based Gaussian Grouping}
\label{subsec:ray_group}

\begin{wrapfigure}{r}{0.50\textwidth}
\centering
\vspace{-10mm}
\includegraphics[width=0.49\textwidth]{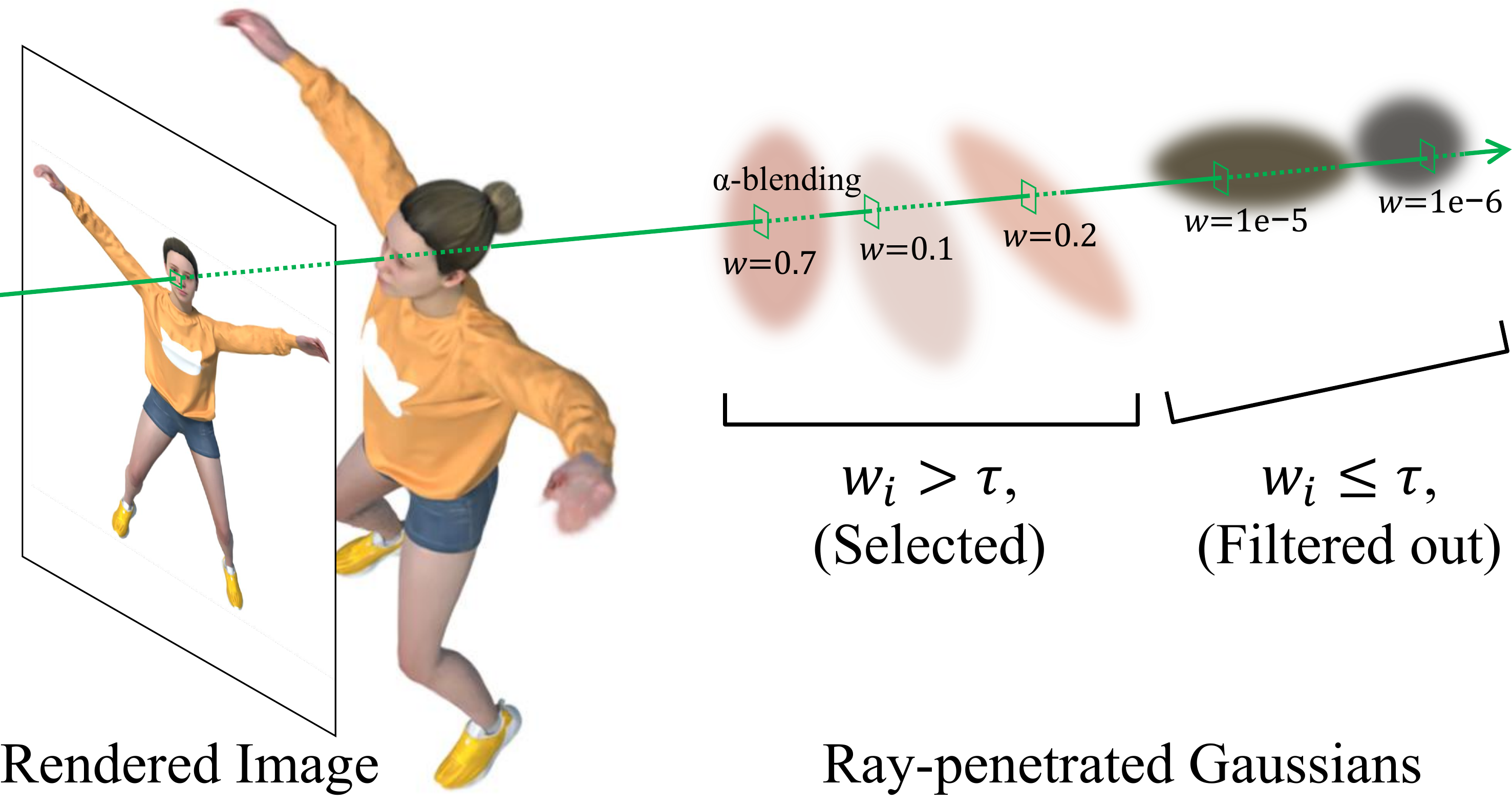}
\vspace{-2mm}
\caption{Ray-based Grouping visualization. We gather the Gaussian penetrated by a ray, and select the Gaussian whose contribution $w_i$ exceeds a threshold $\tau$.}
\label{fig:ray_group}
\vspace{-8mm}
\end{wrapfigure}
Our method begins by selecting Gaussians along each camera ray.
To facilitate object-level motion modeling, we require local groups of primitives that exhibit similar motion.
In 3DGS, the number of primitives changes during training due to adaptive density control, which would require frequent regrouping, making an efficient grouping strategy essential.
To achieve this, we leverage the standard rasterization pipeline, which inherently aggregates and sorts Gaussians along each pixel ray.
We re-purpose this visibility mechanism to construct groups with minimal overhead throughout training.
For each pixel $p_j$, we define the ray-based group $\mathcal{N}_j$ by selecting Gaussians that actively contribute to rendering:
\vspace{-1mm}
\begin{equation}
\mathcal{N}_j = \{ \mathcal{G}_i \mid w_{i} > \tau \},
\label{eq:group_thres}
\end{equation}
where $w_i$ is the blending weight from \cref{eq:alpha_comp}, and $\tau$ is a threshold.
A visualization of grouping is shown in~\cref{fig:ray_group}.
We subsequently impose local regularizers on these groups to encourage coherent motion while preserving local shape structure.

\begin{wrapfigure}{hr}{0.50\textwidth}
\vspace{-3mm}
\centering
\includegraphics[width=0.49\linewidth]{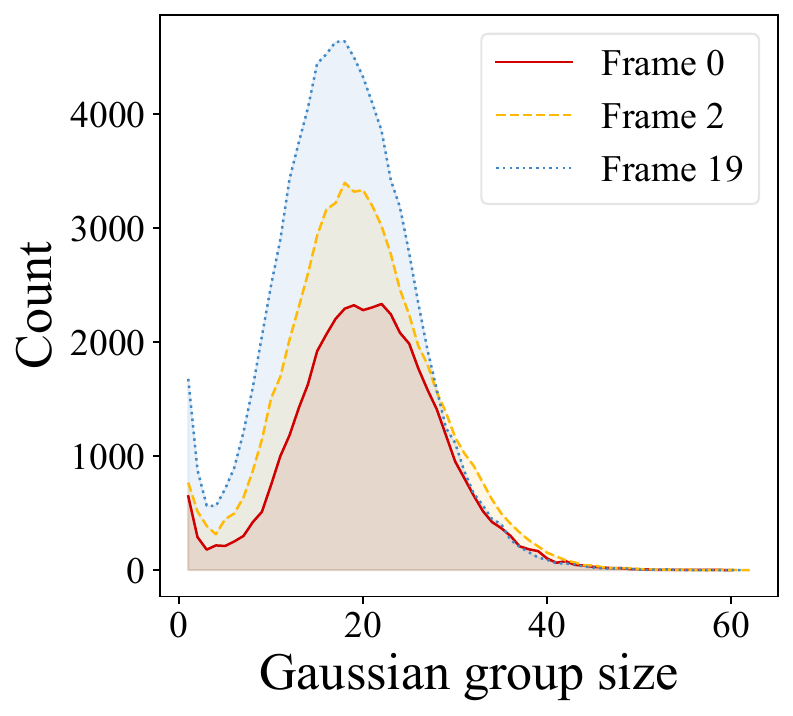}
\hfill
\includegraphics[width=0.49\linewidth]{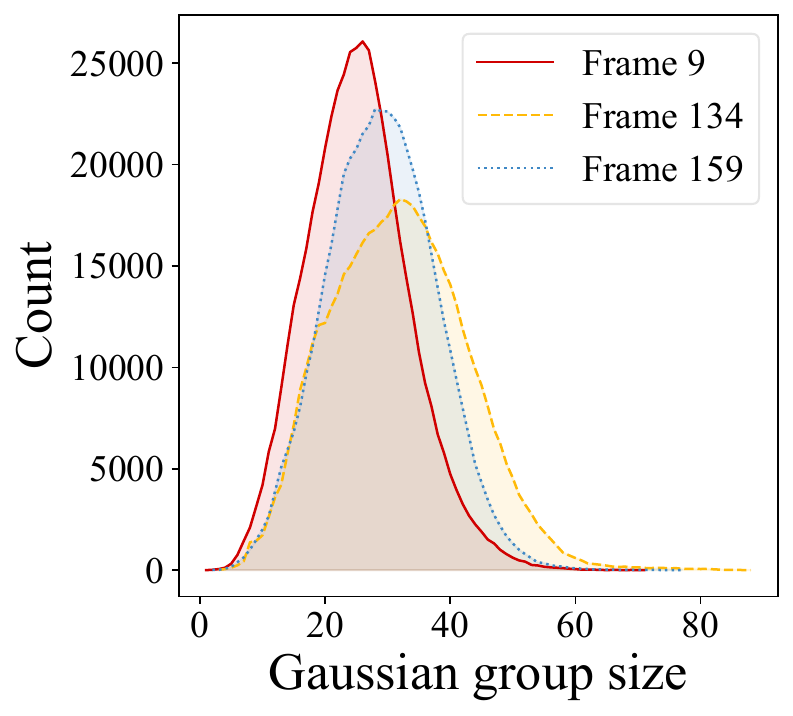}
\caption{Group size visualization. Distribution of Gaussian group sizes for the D-NeRF (left) and HyperNeRF (right). The plot shows the counts of groups with varying sizes across different image frames. Groups of size zero are omitted.}
\label{fig:group_size}
\vspace{-7mm}
\end{wrapfigure}

This design is fundamentally motivated by the physical and geometric properties of the volume rendering process~\cite{mildenhall2020nerf}. The blending weight $w_i$ acts as an implicit, \textbf{occlusion-aware filter}. As the accumulated transmittance attenuates sharply behind opaque occluders~\cite{kerbl20233d}, thresholding by $\tau$ inherently isolates spatially contiguous Gaussians on the unoccluded surface, thereby preventing the entanglement of foreground and background Gaussians. By leveraging this filtered visibility, grouping Gaussians along view rays establishes a direct bridge between 2D observations and 3D geometry. Instead of relying on explicit 3D distance metrics that might erroneously cluster physically close but structurally independent parts~\cite{qi2017pointnetplusplus}, our ray-based grouping establishes a dense set of spatio-temporal constraints to robustly regularize object motion. Furthermore, this formulation naturally yields highly flexible and adaptive group sizes. As demonstrated by the wide distribution in~\cref{fig:group_size}, our approach dynamically accommodates varying local complexities, ranging from thin structures to dense volumes, without the need for explicit structural priors or heuristic guidance.

\vspace{-2mm}
\subsection{Motion Coherence Regularization}
\label{subsec:cov_reg}
We first introduce a motion coherence regularization (MCR) term that encourages coherent dynamics within each ray-based group $\mathcal{N}_j$ without relying on external supervision such as optical flow or monocular depth~\cite{zhu2024motiongs}.
Instead, we regularize directional consistency while allowing magnitudes to adapt.
Let $\boldsymbol{\mu}_{i,t}$ be the 3D position of Gaussian $\mathcal{G}_i \in \mathcal{N}_j$ at time $t$.
For a temporal offset $\Delta t$, the displacement is:
\begin{equation}
\vd_{i,t} = \boldsymbol{\mu}_{i,t+\Delta t} - \boldsymbol{\mu}_{i,t}.
\end{equation}
We compute the mean displacement of a group:
\begin{equation}
\bar{\vd}_{\mathcal{N}_j,t}
= \frac{1}{|\mathcal{N}_j|} \sum_{\mathcal{G}_i \in \mathcal{N}_j} \vd_{i,t}. 
\end{equation}
We then penalize directional inconsistency via a cosine-similarity loss:
\begin{equation}
\mathcal{L}_{\mathrm{MCR}}
= 1 - \frac{ \vd_{i,t} \cdot \bar{\vd}_{\mathcal{N}_j,t} }
{\lVert \vd_{i,t}\rVert \lVert \bar{\vd}_{\mathcal{N}_j,t} \rVert + \epsilon },
\end{equation}
where $\epsilon$ is a small positive number. The loss is applied only to moving Gaussians satisfying $\lVert\vd_{i,t}\rVert > 0.0001$.
This regularization encourages Gaussians that contribute to the same pixel to move coherently, while still permitting spatially varying motion magnitudes. Crucially, we do not penalize differences in displacement magnitudes. This relaxed constraint is essential, as enforcing uniform motion magnitudes would erroneously impose strict rigid translations by conflating coherent motion with identical displacement.

\vspace{-2mm}
\subsection{Spectral Regularization}
\label{subsec:welford}
MCR effectively aligns the overall trajectories of Gaussians, but lacks an explicit mechanism to preserve local spatial structures over time. A common geometric prior for this task is local rigidity, typically enforced through ARAP energy functions~\cite{sorkine2007arap, huang2023sc} that penalize changes in pairwise distances and relative orientations. However, enforcing such point-to-point rigidity is overly restrictive for dynamic scenes, as it severely hinders natural non-rigid deformations and smooth motion.

Rather than imposing strict point correspondences, we propose a spectral regularization (SR) that maintains the distributional shape of the group. We compute the covariance matrix of Gaussian positions at time steps $t$ and $t+\Delta t$ and penalize the difference between their eigenvalue spectra. 
Let $K_t$ and $K_{t+\Delta t}$ be covariances of the same ray group at $t$ and $t+\Delta t$.
Let $\sigma_{t,r}$ and $\sigma_{t+\Delta t,r}$ be the $r$-th eigenvalues of $K_t$ and $K_{t+\Delta t}$ in ascending order, respectively.
We define the covariance consistency loss as follows:
\begin{equation}
\mathcal{L}_{\mathrm{SR}}   
= \sum_{r \in \{1,2,3\}} \mathrm{Huber}\bigl(\sigma_{t,r}, \sigma_{t+\Delta t,r}\bigr),
\end{equation}
where $\mathrm{Huber}(\cdot,\cdot)$ is the standard Huber loss~\cite{huber64}. We apply SR only to groups with more than one Gaussian.

Matching eigenvalue spectra preserves local shape statistics over time, while remaining invariant to rigid rotations and allowing flexible non-rigid motions. This preserves the spatial volume while allowing localized deformations. This formulation prevents shape distortion even when the group forms a broad spatial distribution. Since the loss function does not minimize the overall scale of the eigenvalue spectrum, the regularizer does not force the group to contract even if it contains distant Gaussians. Instead, it selectively targets Gaussians that exhibit displacements large enough to alter the object shape. The regularization guides these deviating Gaussians to maintain spatial proximity with their grouped neighbors, restricting only the movements that disrupt the structure.

\vspace{4pt}
\paragrapht{ARAP baseline for comparison.}
\label{sec:ARAP}
For completeness, we additionally consider an ARAP-style rigidity prior~\cite{sorkine2007arap,huang2023sc} as a \emph{comparative} alternative to $\mathcal{L}_{\mathrm{SR}}$ in our experiments.
Concretely, for each Gaussian $i$ with neighbors $\mathcal{M}(i)$, we estimate a local rotation $\bm{R}_{i,t}$ that best aligns relative vectors across time and penalize residual distortion:
\begin{equation}
\mathcal{L}_{\mathrm{ARAP}}
= \sum_{i} \sum_{j \in \mathcal{M}(i)}
\left\|
(\boldsymbol{\mu}_{i,t+\Delta t}-\boldsymbol{\mu}_{j,t+\Delta t})
- \bm{R}_{i,t}\,(\boldsymbol{\mu}_{i,t}-\boldsymbol{\mu}_{j,t})
\right\|_2^2.
\end{equation}
We use this term only to contextualize the effect of strict local rigidity relative to our spectral regularization.

\subsection{Welford's Algorithm}
We employ Welford's algorithm~\cite{Welford1962Note} for efficient, online covariance calculation. Consider a group consisting of $N$ Gaussians. Let $K_{i-1}$ represent the covariance matrix computed from the first $i-1$ Gaussian positions. When incorporating the $i$-th Gaussian, the covariance $K_i$ is updated recursively as follows:
\begin{equation}
K_i = \frac{i-1}{i} K_{i-1} + \frac{1}{i} P_i,
\quad
P_i := \bigl(\boldsymbol{\mu}_i - \bar{\boldsymbol{\mu}}_{i-1}\bigr)
       \bigl(\boldsymbol{\mu}_i - \bar{\boldsymbol{\mu}}_i\bigr)^{\top},
\quad i = 1,\dots,N,
\end{equation}
where $\bar{\boldsymbol{\mu}}_i = \frac{1}{i}\sum_{j=1}^{i} \boldsymbol{\mu}_j$ is the running mean.
By induction,
\begin{equation}
K_N
= \sum_{i=1}^N
\left(
\frac{1}{i}
\prod_{k=i+1}^N \frac{k-1}{k}
\right) P_i
= \frac{1}{N} \sum_{i=1}^N P_i.
\end{equation}
This enables covariance computation along each ray in a single pass and can be integrated into the rasterization pipeline (a proof in~\cref{sec:welford_proof}).

\subsection{Final Objective}
We adopt the photometric losses used in 3DGS, namely the $\ell_1$ loss $\mathcal{L}_1$ and the structural dissimilarity loss $\mathcal{L}_{\mathrm{dssim}}$, and augment them with our regularization terms.
Our default training objective is:
\begin{equation}
    \mathcal{L}
    = (1-\lambda_{\mathrm{dssim}})\mathcal{L}_{1}
    + \lambda_{\mathrm{dssim}}\mathcal{L}_{\mathrm{dssim}}
    + \lambda_{\mathrm{MCR}} \, \mathcal{L}_{\mathrm{MCR}}
    + \lambda_{\mathrm{SR}} \, \mathcal{L}_{\mathrm{SR}},
\end{equation}
where $\lambda_{\mathrm{dssim}}$, $\lambda_{\mathrm{MCR}}$, and $\lambda_{\mathrm{SR}}$ are the hyperparameters.
In the ARAP comparison setting, we replace $\mathcal{L}_{\mathrm{SR}}$ with $\mathcal{L}_{\mathrm{ARAP}}$ (with its own weight) and set $\lambda_{MCR}=0$ while keeping all other components identical.

%% file: sections/4_experiment.tex
\vspace{2mm}
\section{Experiment}

\vspace{2mm}
\subsection{Datasets}
\label{sec:datasets}
We evaluate our method on the D-NeRF~\cite{pumarola2020dnerf}, HyperNeRF~\cite{park2021hypernerf}, and NeRF-DS~\cite{yan2023nerfds} datasets. Following prior protocols, we report PSNR, SSIM, and VGG-based LPIPS (LPIPS$_{V}$)~\cite{zhang2018perceptual} for D-NeRF; PSNR, MS-SSIM (MS), and AlexNet-based LPIPS (LPIPS$_{A}$) for HyperNeRF; and PSNR, SSIM, MS-SSIM, and VGG-based LPIPS for NeRF-DS.

\vspace{5pt}
\noindent\textbf{D-NeRF.} \quad
D-NeRF contains eight synthetic scenes featuring object deformations, captured using a 360$^{\circ}$ monocular camera setup. Following prior works~\cite{pumarola2020dnerf,wu20234dgs}, we perform evaluation at a resolution of $800 \times 800$. 
We exclude the \textit{Lego} scene because the training and test splits are not temporally aligned~\cite{yang2023deformable,xu2024grid4d}.

\vspace{5pt}
\noindent\textbf{HyperNeRF.} \quad
HyperNeRF features real-world scenes including noise, illumination variations, and topological changes. We use the vrig split, which is captured with a handheld dual-camera rig that simultaneously records two views per frame during data collection. The training set is constructed by alternately selecting images from the left and right cameras for each frame, with the remaining images reserved for testing.
For all experiments, images are downsampled to half their original resolution, \ie, $536 \times 960$. Camera poses are estimated using COLMAP~\cite{schonberger2016structure}, and point clouds are obtained from RTD~\cite{wu20234dgs}.

\vspace{5pt}
\noindent\textbf{NeRF-DS.} \quad
NeRF-DS comprises seven real-world scenes with specular objects and is provided at a resolution of $480 \times 270$. The dataset is captured with a dual-camera setup, using the left camera for training and the right-camera for testing.

\vspace{-2mm}
\subsection{Baseline models}
\vspace{-1mm}
We compare our method against four representative baselines: RTD~\cite{wu20234dgs}, MoDec-GS~\cite{kwak2025modecgs}, Grid4D~\cite{xu2024grid4d}, and Ex4DGS~\cite{lee2024fully}. RTD, MoDec-GS, and Grid4D adopt deformation-based formulations, whereas Ex4DGS uses spline parameterization.
Specifically, RTD parameterizes the deformation field with HexPlanes~\cite{cao2023hexplane}, MoDec-GS employs scaffold-based representations~\cite{Lu2024scaffoldgs} to model motion, and Grid4D utilizes a hash-grid representation~\cite{mueller2022instant} for deformation encoding. In contrast, Ex4DGS represents dynamic Gaussians explicitly through spline trajectories.

\vspace{5pt}
\noindent\textbf{Implementation.} \quad
All experiments are conducted on a single NVIDIA RTX 3090 GPU. We build upon the official implementations of the baseline methods and adopt the dataset-specific parameter settings provided in their respective protocols. For fair comparison, we keep the baseline architectures and hyperparameters unchanged, except for the additional regularization introduced by our method. Detailed implementation specifics are provided in \cref{sec:impl_details}.

\input{tables/main_comp}

\begin{figure}[t]
\centering
 \includegraphics[width=0.95\columnwidth, page=1]{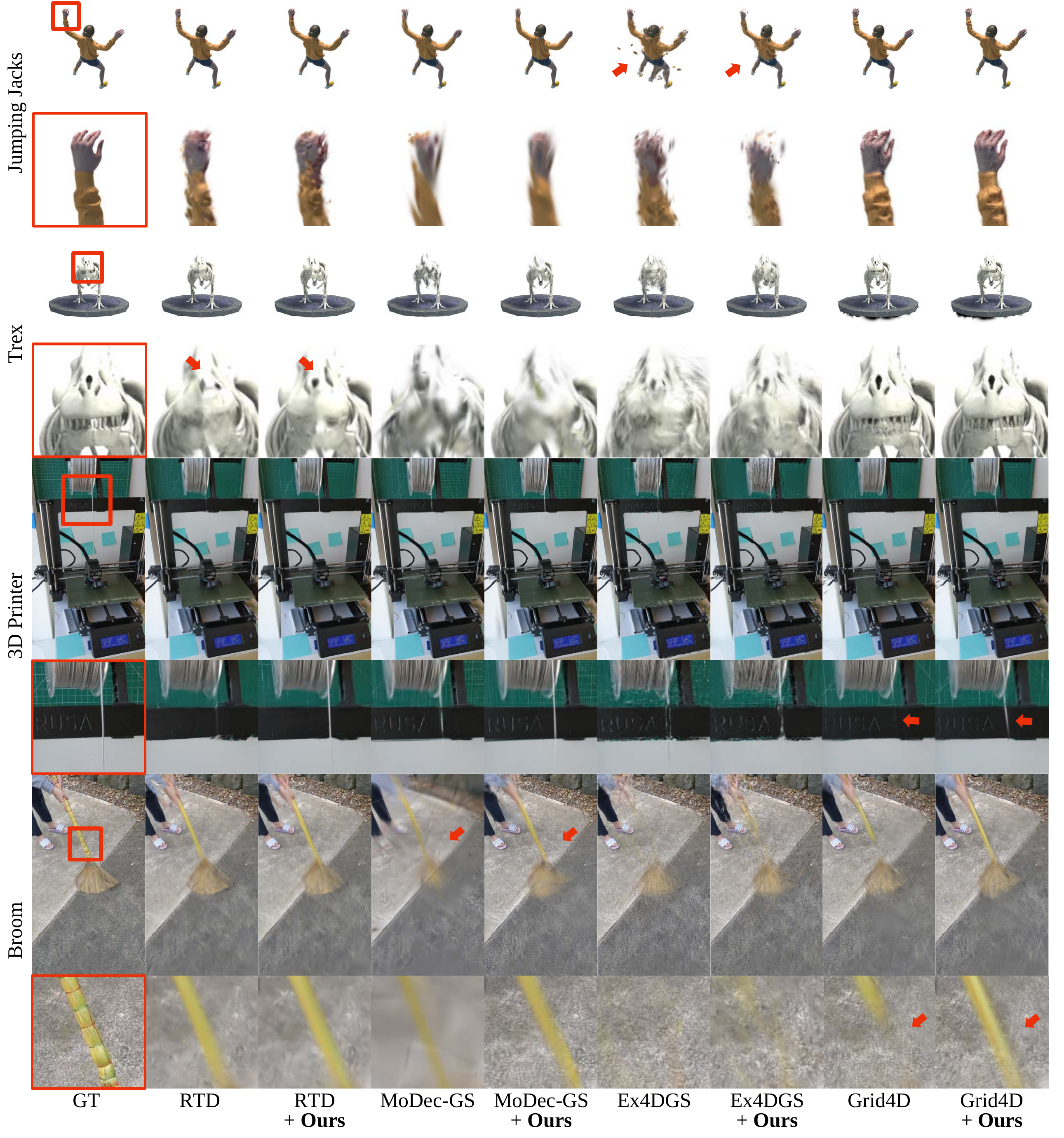}
 \vspace{-3mm}
 \caption{Qualitative comparisons between baselines and our method on the D-NeRF and HyperNeRF datasets.}
 \label{fig:qual_comp} 
\vspace{-4mm}
\end{figure}

\subsection{Comparison}
\label{sec:comparison}
\subsubsection{Quantitative comparison.}

\cref{tab:performance_comparison,tab:nerfds_comp} report quantitative results on the D-NeRF, HyperNeRF, and NeRF-DS. Our method consistently improves reconstruction quality when integrated with all baselines. On the D-NeRF, we observe clear PSNR gains across all models, \eg, $+1.11$ dB for Ex4DGS and $+2.35$ dB for MoDec-GS, while Grid4D$+$Ours achieves the best PSNR of $42.20$. On the HyperNeRF, our regularization improves both PSNR and SSIM for most baselines, with Grid4D$+$Ours obtaining the highest MS-SSIM of $0.856$. Notably, on the more challenging NeRF-DS dataset, our method provides larger improvements, particularly for MoDec-GS and Grid4D, leading to significant gains in both PSNR and perceptual quality. These results indicate that our regularization is complementary to existing dynamic 3DGS methods and consistently enhances reconstruction performance across diverse datasets.

\begin{figure}[t]
\centering
 \includegraphics[width=1.\columnwidth, page=2]{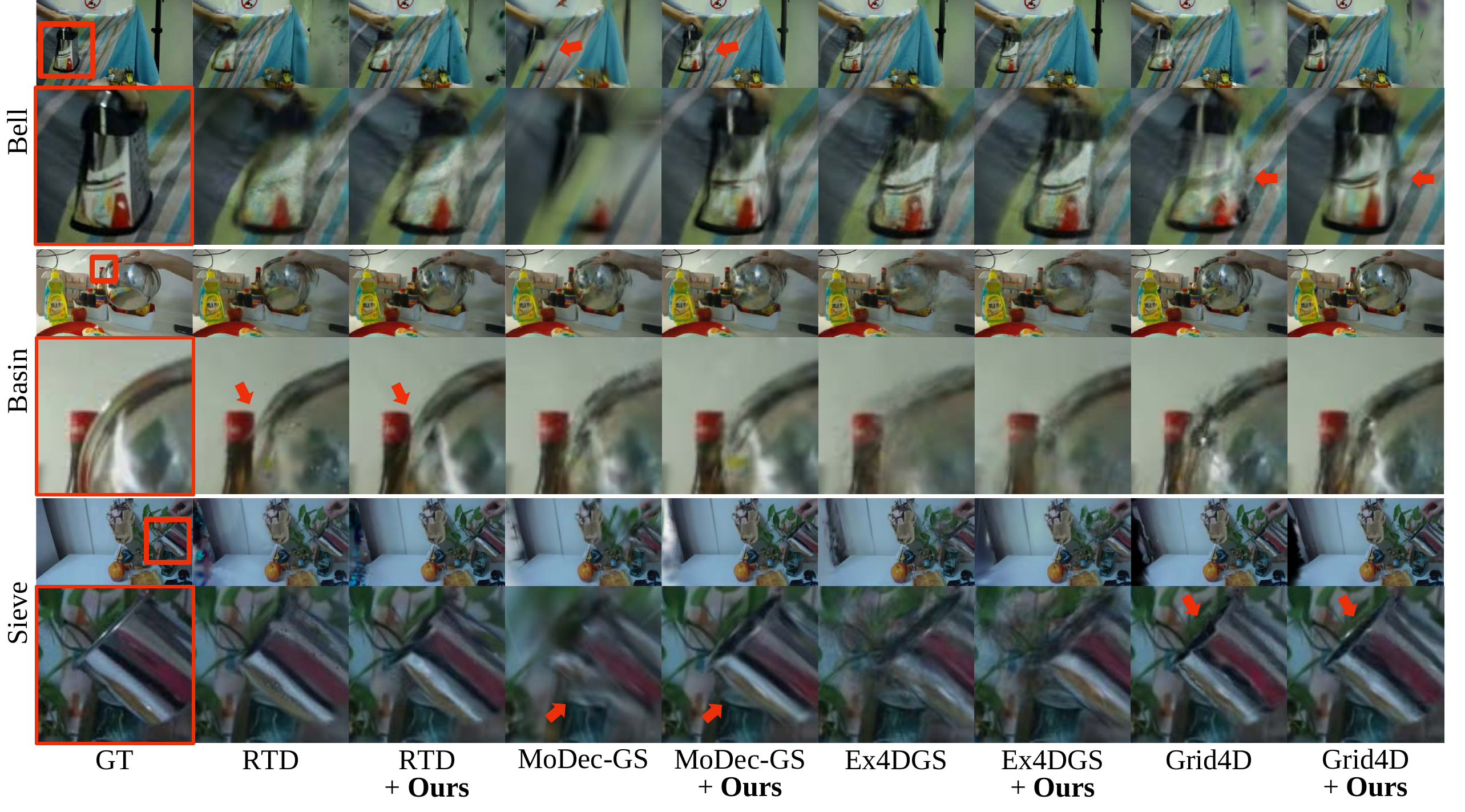}
 \vspace{-7mm}
 \caption{Qualitative comparisons between baselines and our method on the NeRF-DS dataset.}
 \label{fig:qual_comp_nerfds}
\vspace{-5mm}
\end{figure}

\input{tables/nerfds}

\subsubsection{Qualitative comparison.}
\cref{fig:qual_comp,fig:qual_comp_nerfds} present visual comparisons on the D-NeRF, HyperNeRF, and NeRF-DS datasets. Existing methods produce artifacts such as disappearing objects or distorted shapes, whereas our method preserves the structural integrity of Gaussians over time. 
On the HyperNeRF, prior methods tend to blur or remove thin structures, like the handle of the broom, while our regularization preserves structures and maintains physically plausible shapes. Similarly, in the \textit{JumpingJacks} scene of the D-NeRF, our method better retains fine details compared to the baselines. This improvement arises because our regularization aligns Gaussian motions with the underlying object dynamics, preventing incorrect motion updates during image-based optimization.
Furthermore, as shown in \cref{fig:qual_comp}, our ray-based grouping strategy enables reliable grouping for small structures, like the fingers in \textit{JumpingJacks} scene and the teeth in \textit{Trex} scene, facilitating accurate shape optimization for methods like Grid4D. On the NeRF-DS \textit{Bell} scene, despite the presence of specular reflections, enforcing physically consistent motion leads to improved reconstruction quality. 

\subsubsection{Ablation study.}
\input{tables/RTD_ablation}

\begin{figure}[t]
\centering
 \includegraphics[width=0.65\columnwidth]{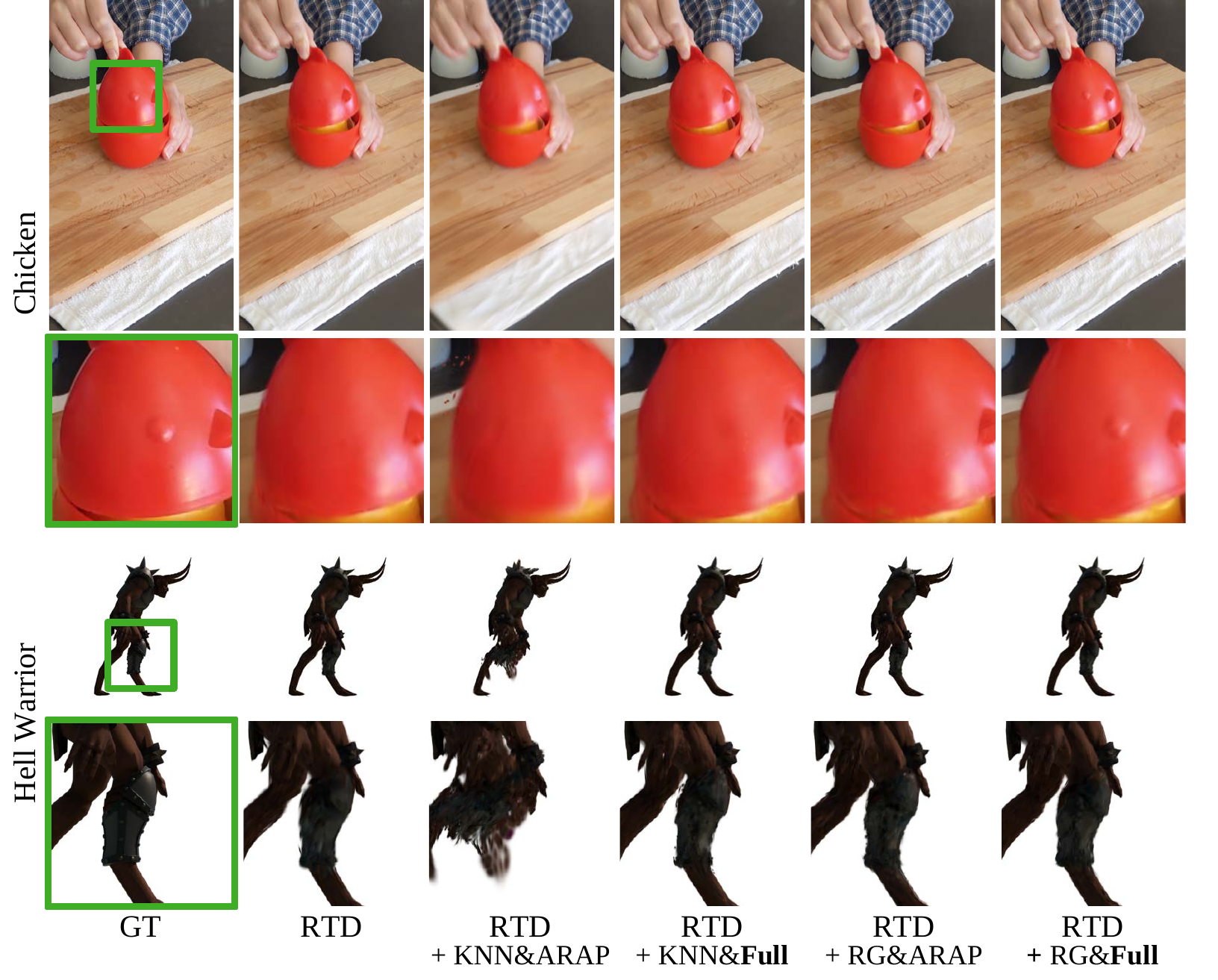}
 \vspace{-1mm}
 \caption{Ablation results on the RTD model. We compare KNN grouping and ray-based grouping (RG) under different regularization settings, including ARAP, MCR, and SR. \textbf{Full} denotes the configuration where both MCR and SR are applied.}
 \label{fig:ablation}
\vspace{-3mm}
\end{figure}

\cref{tab:abl_comparison,fig:ablation} present ablation studies on the D-NeRF and HyperNeRF datasets using RTD as the base model. We report the average results over all scenes in each dataset. Specifically, we compare our ray-based grouping (RG) with a KNN grouping strategy and analyze the effect of removing each component, including ARAP, MCR, and SR. Except for the introduced components, all hyperparameters remain identical to the baseline configuration. We additionally report the training time to analyze the computational overhead. 
For the KNN baseline, we follow the strategy used in D3DGS~\cite{luiten2023dynamic}, which constructs a group by selecting the 20 nearest Gaussians at every iteration. For ARAP regularization, we adopt the formulation from~\cref{sec:ARAP}. The \textbf{Full} configuration denotes the model with both MCR and SR enabled. Detailed numerical results are provided in~\cref{sec:abl_impl_details}.
\paragraph{KNN comparison.}
We first compare our ray-based grouping with the KNN grouping strategy. Since KNN grouping does not account for Gaussian scale or orientation, it often fails to select physically coherent groups. As shown in~\cref{tab:abl_comparison}, this leads to degraded performance for both ARAP and \textbf{Full} configurations. Incorrect grouping causes unrelated Gaussians to move together, resulting in inconsistent motion and reduced reconstruction quality.

The qualitative results illustrate this issue. As shown in~\cref{fig:ablation}, KNN grouping frequently produces mis-grouped Gaussians. In the \textit{Hell Warrior} scene, applying strong rigidity constraints such as ARAP to incorrectly grouped Gaussians can even cause parts of the object to disappear. In contrast, our ray-based grouping preserves the structural consistency of the object. Moreover, the visibility-based thresholding prevents grouping across occluded regions and helps preserve fine details, such as the eyes of the chicken in the RG+\textbf{Full} configuration.

\paragraph{ARAP comparison.}
We further evaluate ARAP regularization with both KNN and RG grouping strategies, as reported in \cref{tab:abl_comparison}. When combined with RG, ARAP produces comparable or improved results compared to KNN. However, on the D-NeRF dataset, the performance still decreases. This is because the rigid transformation assumption in ARAP restricts fine-grained shape variations, preventing the model from capturing detailed geometric changes.
We also analyze the individual effects of MCR and SR. MCR encourages directional consistency among neighboring Gaussians but does not explicitly preserve shape. Thus, it works well for scenes with relatively linear motion, such as those in the D-NeRF dataset, but becomes less effective in the more complex dynamics of HyperNeRF. In contrast, SR focuses solely on shape preservation and does not constrain motion, which makes it insufficient when applied alone. These results indicate that both a well-defined grouping strategy and an appropriate motion prior are necessary to achieve consistent performance improvements.

\paragraph{Training time.}
The training time overhead introduced by the proposed regularization is reported in~\cref{tab:abl_comparison}. The increase is more noticeable on the HyperNeRF dataset, which contains a larger number of Gaussians and higher pixel counts. Most of the additional cost arises during rasterization, particularly from covariance processing and SVD operations required for the regularization terms. Despite this overhead, RG remains $6\%$–$25\%$ faster than KNN in training time under the \textbf{Full} configuration.

\begin{figure}[t]
\centering
 \includegraphics[width=1.0\linewidth]{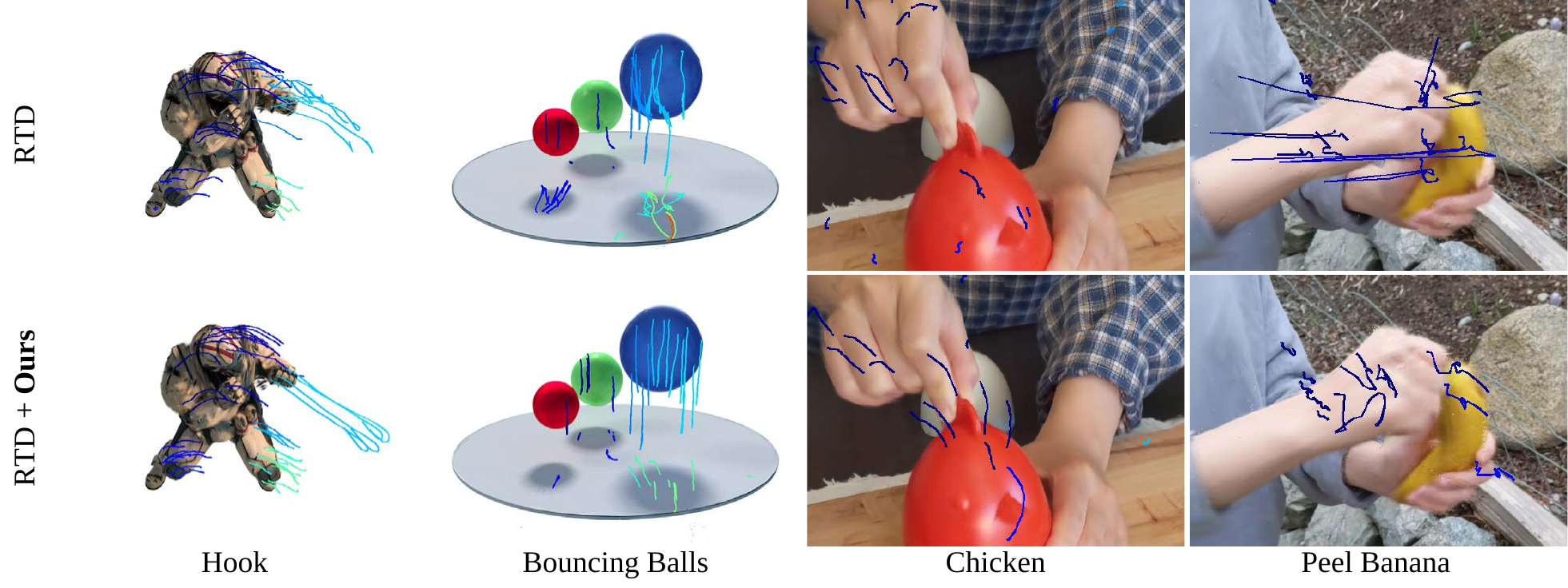}
 \vspace{-6mm}
 \caption{Gaussian trajectory visualization results of the RTD with our method on the D-NeRF and HyperNeRF datasets. We project the temporal positions of sampled Gaussians onto a fixed camera view.}
 \label{fig:traj_vis}
\vspace{-6mm}
\end{figure}

\subsubsection{Trajectory visualization.}
\cref{fig:traj_vis} visualizes Gaussian trajectories of the RTD model for both the baseline and our method on the D-NeRF and HyperNeRF datasets. We visualize the trajectories of Gaussian positions by projecting them onto a fixed camera to analyze their motion.
On the D-NeRF, our method produces trajectories that are better aligned with the underlying physical motion compared to the baseline. While the baseline often exhibits permuted or irregular Gaussian paths, our method results in more structured and temporally consistent trajectories. This behavior arises because our regularization encourages neighboring Gaussians to move coherently.
On the HyperNeRF dataset, particularly around the hand region, the baseline often produces Gaussians that drift away from the object surface. In contrast, our method keeps Gaussians consistently attached to the same object region. This result indicates that our approach effectively regularizes outlier Gaussians and improves motion consistency.

%% file: tables/main_comp.tex
\begin{table*}[!t]
\centering
\caption{Quantitative comparison of our method on the D-NeRF and HyperNeRF dataset. We report PSNR, MS-SSIM, and LPIPS (AlexNet). $\Delta$ denotes the PSNR improvement when the baseline models are combined with our framework.}
\label{tab:performance_comparison}
\resizebox{0.8\linewidth}{!}
{%
\begin{tabular}{l c cccc c cccc}
\toprule
\multirow{2}{*}{\centering Model} & ~~~ & \multicolumn{4}{c}{D-NeRF} & ~~ & \multicolumn{4}{c}{HyperNeRF} \\
\cmidrule(){3-6} \cmidrule(){8-11}
& &\! PSNR$\uparrow$ ~ & \!\!SSIM$\uparrow$\!\! \, & LPIPS$_{V}\!\downarrow$\! & $\Delta\!\uparrow$ & & \! PSNR$\uparrow$ ~ & \!\!\!MS-SSIM$\uparrow$\!\!\! \, & LPIPS$_{A}\!\downarrow$\! & $\Delta\!\uparrow$ \\
\midrule
NeRF~\cite{mildenhall2020nerf} & & 18.81 & 0.881 & 0.176 & - & & 20.13 & 0.745 & 0.424 & - \\
D-NeRF~\cite{pumarola2020dnerf} & & 31.69 & 0.970 & 0.043 & - & & \NA & \NA & \NA & -  \\
TiNeuVox~\cite{fang2022TiNeuVox} & & 33.76 & 0.980 & 0.039 & - & & 24.25 & 0.837 & \NA & - \\
HyperNeRF~\cite{park2021hypernerf} & & \NA & \NA & \NA & - & & 22.38 & 0.814 & \NA & - \\
3D-GS~\cite{kerbl20233d} & & 23.39 & 0.929 & 0.077 & -  & & 20.26 & 0.657 & 0.342 & - \\
D3DGS~\cite{yang2023deformable} & & 40.43 & 0.992 & 0.011 & - & & 22.40 & \NA & 0.275 & - \\
GaGS~\cite{lu20243d} & & 39.07 & \tul{0.993} & \tbf{0.007} & - & & 24.26 & 0.793 & \NA & - \\
\midrule
Ex4DGS~\cite{lee2024fully} & & 29.93 & 0.968 & 0.041 & - & & 24.61 & 0.829 & 0.199 & -\\
RTD~\cite{wu20234dgs} & & 35.36 & 0.985 & 0.022 & - & & 25.17 & 0.841 & 0.283 & - \\
MoDec-GS~\cite{kwak2025modecgs} & & 29.33 & 0.963 & 0.048 & - & & 24.92 & 0.841 & \tul{0.176} & - \\
Grid4D~\cite{xu2024grid4d} & & \tul{42.00} & \tbf{0.994} & \tul{0.008} & - & & \tul{25.50} & 0.783 & 0.185 & - \\ \midrule
Ex4DGS$+$\tbf{Ours} & & 31.04 & 0.972 & 0.038 & 1.11 & & 24.86 & 0.836 & 0.233 & 0.25 \\
RTD$+$\tbf{Ours} & & 36.46 & 0.987 &0.019 & 1.10 & & 25.30 & 0.845 & 0.273 & 0.13 \\
MoDec-GS$+$\tbf{Ours} & & 31.68 & 0.974 & 0.035 & 2.35 & & 25.09 & \tul{0.846} & \tbf{0.173} & 0.17 \\
Grid4D$+$\tbf{Ours} & & \tbf{42.20} & \tbf{0.994} & \tul{0.008} & 0.20 & & \tbf{25.56} & \tbf{0.856} & 0.198 & 0.06 \\
\bottomrule
\end{tabular}%
}
\end{table*}

%% file: tables/nerfds.tex
\begin{table*}[ht]
\centering
\caption{Quantitative comparison on the NeRF-DS dataset. We report PSNR, SSIM, MS-SSIM and LPIPS (VGG). $\Delta$ denotes the PSNR improvement when the baseline models are combined with our framework.}
\vspace{-3mm}
\label{tab:nerfds_comp}
\resizebox{0.68\linewidth}{!}{%
\setlength{\tabcolsep}{4pt}
\begin{tabular}{l c cccc}
\toprule
\multirow{2}{*}{\centering Model} & ~~~ & \multicolumn{4}{c}{NeRF-DS} \\
\cmidrule(){3-6} 
& & PSNR$\uparrow$ & \! SSIM/MS$\uparrow$ & \!\!\! LPIPS$_{V}\downarrow$ \! & $\Delta\!\uparrow$ \\
\midrule
TiNeuVox~\cite{fang2022TiNeuVox}    & & 21.61       & 0.823/~~~-~~~ & 0.277 & -  \\
HyperNeRF~\cite{park2021hypernerf}  & & 23.45       & ~~~-~~~/0.849 & 0.199 & -  \\
NeRF-DS~\cite{yan2023nerfds}        & & 23.60       & ~~~-~~~/0.849 & \tbf{0.182} & -  \\
3D-GS~\cite{kerbl20233d}            & & 20.29       & 0.782/~~~-~~~ & 0.292 & - \\
D3DGS~\cite{yang2023deformable}     & & 23.61       & 0.839/~~~-~~~ & 0.197 & - \\ \midrule
Ex4DGS~\cite{lee2024fully}          & & 21.73       & 0.793/0.789   & 0.251 & - \\
RTD~\cite{wu20234dgs}               & & 21.91       & 0.812/0.821   & 0.235 & - \\
MoDec-GS~\cite{kwak2025modecgs}     & & 22.93       & 0.822/0.851   & 0.229 & - \\
Grid4D~\cite{xu2024grid4d}          & & 23.41       & 0.833/0.872   & 0.196 & - \\ \midrule
Ex4DGS+\tbf{Ours}                   & & 21.88       & 0.798/0.794   & 0.248 & 0.15 \\
RTD+\tbf{Ours}                      & & 22.05       & 0.813/0.826   & 0.226 & 0.14 \\
MoDec-GS+\tbf{Ours}                 & & \tbf{23.76} & \tbf{0.848}/\tbf{0.885} & \tul{0.184} & 0.83 \\
Grid4D+\tbf{Ours}                   & & \tul{23.70} & \tul{0.840}/\tul{0.882} & 0.188 & 0.32 \\
\bottomrule\end{tabular}%
}
\end{table*}

%% file: tables/RTD_ablation.tex
\begin{table}[t]
    \centering
    \caption{Ablation comparison table. Experiments are conducted on the RTD model, and we report the average results on the D-NeRF and HyperNeRF datasets. KNN denotes the use of KNN-based grouping, and ARAP indicates as-rigid-as-possible regularization. RG represents ray-based grouping, MCR denotes motion coherence regularization, and SR denotes spectral regularization. \textbf{Full} indicates the configuration where both MCR and SR are applied. Time refers to the training time.}
    \vspace{-1mm}
    \label{tab:abl_comparison}
    \setlength{\tabcolsep}{3pt}
    \resizebox{0.95\linewidth}{!}{%
    \begin{tabular}{l c cccc c cccc}
        \toprule
        \multirow{2}{*}{\centering Model} & ~~~ & \multicolumn{4}{c}{D-NeRF} & ~~ & \multicolumn{4}{c}{HyperNeRF} \\
        \cmidrule(){3-6} \cmidrule(){8-11}
        &&  PSNR$\uparrow$ & SSIM$\uparrow$ & LPIPS$_{V}\downarrow$ & Time(s) & & PSNR$\uparrow$ & MS-SSIM$\uparrow$ & LPIPS$_{A}\downarrow$ & Time(s) \\
        \midrule
        RTD~\cite{wu20234dgs}    & & 35.36 & 0.9845 & 0.0224 & \tbf{841}  & & 25.17 & 0.8413 & 0.2827 & \tbf{1963}  \\
        RTD$+$KNN\&ARAP          & & 34.80 & 0.9828 & 0.0258 & \tul{1484} & & 23.18 & 0.7532 & 0.4606 & 5808  \\
        RTD$+$KNN\&\textbf{Full} & & 33.29 & 0.9771 & 0.0290 & 2044 & & 25.11 & 0.8390 & 0.2856 & 9586  \\ \midrule
        RTD$+$RG\&ARAP           & & 35.27 & 0.9844 & 0.0216 & 2125 & & \tul{25.25} & 0.8430 & \tul{0.2710} & 13096  \\
        RTD$+$RG\&MCR            & & \tul{36.23}& \tul{0.9866} & \tbf{0.0190} & 1649 & & 24.23 & \tul{0.8449} & 0.2713 & \tul{4454}  \\
        RTD$+$RG\&SR             & & 34.98 & 0.9841 & 0.0220 & 1501 & & 24.91 & 0.8331 & 0.2961 & 6694  \\
        RTD$+$RG\&\textbf{Full}  & & \tbf{36.46} & \tbf{0.9869} & \tul{0.0193} & 1924 & & \tbf{25.30} & \tbf{0.8564} & \tbf{0.2488} & 7138  \\
        \bottomrule
    \end{tabular}
    }
\vspace{-2mm}
\end{table}

%% file: sections/5_conclusion.tex
\section{Conclusion}
Modeling physically plausible motion in dynamic 3D Gaussian Splatting remains challenging, as many existing approaches rely heavily on external priors to mitigate physically implausible motion. In this work, we propose a method that enables dynamic 3DGS models to learn physically plausible motion directly from image supervision. By introducing a ray-based grouping strategy coupled with motion coherence and spectral regularization, our approach effectively enforces relaxed rigidity constraints. We integrate our approach into four baseline models and evaluate them across three datasets, consistently observing performance improvements. Notably, integrating our method with Grid4D yields state-of-the-art results, demonstrating both the effectiveness and generality of our framework. Our findings confirm that physically grounded motion constraints can significantly improve dynamic 3DGS reconstruction without relying on external priors. Future work will explore finer-grained physical constraints to facilitate robust modeling under highly complex or restricted motion conditions.

%% file: sections/X_suppl.tex
\clearpage
\setcounter{page}{1}
\maketitlesupplementary

\section{Supplementary Overview}

This supplementary material is organized as follows. \cref{sec:impl_details} provides in-depth details on the model implementation and hyperparameter settings. \cref{sec:abl_impl_details} presents the detailed setup of the ablation study. \cref{sec:welford_proof} describes the proposed formulation of Welford's algorithm for the Gaussian splatting. Finally, \cref{sec:additional_vis} reports details of our per-scene quantitative results. The \textit{demo video} and the \textit{source code} of our model are included in our \textit{supplementary material}.

\section{Implementation Details}
\label{sec:impl_details}

In this section, we describe the implementation of the proposed ray-based grouping, Motion Coherence Regularization (MCR), and Spectral Regularization (SR), together with their associated hyperparameters.

Our ray-based grouping is implemented on top of the 3DGS rasterizer. Specifically, we add a module to the rendering pipeline that filters Gaussians within each group by applying a threshold $\tau$ to their $\alpha$-blending weights, so that only the selected Gaussians are processed for regularization. Within this module, we compute the average displacement of grouped Gaussians and their covariance matrix. The covariance computation follows the formulation in~\cref{sec:welford_proof}.

The averaged displacement is used to compute the MCR loss, while the SR loss is derived from the eigenvalues obtained by applying singular value decomposition (SVD) to the covariance matrix. To estimate temporal motion, we uniformly sample a timestep from the interval $[t-m,\,t+m]$ centered at the current timestep $t$, where $m$ controls the temporal sampling range.

We follow the original training configuration of each baseline model. For the NeRF-DS, we adopt the hyperparameters based on the HyperNeRF configuration. For RTD on the HyperNeRF, we additionally increase the number of training iterations to 20,000. Across datasets, the temporal interval parameter $m$ is selected from $[2, 20]$ for the D-NeRF, $[4, 20]$ for the HyperNeRF, and $[2, 16]$ for the NeRF-DS. The regularization weight $\lambda_{MCR}$ is chosen from $[10^{-4}, 10^{-2}]$ for the D-NeRF, $[10^{-3}, 5\times10^{-2}]$ for the HyperNeRF, and $[10^{-3}, 10^{-2}]$ for the NeRF-DS. The structural regularization weight $\lambda_{SR}$ is chosen from $[10^{-3}, 10^{-1}]$ for the D-NeRF, $[10^{-4}, 1]$ for the HyperNeRF, and $[10^{-2}, 10^{-1}]$ for the NeRF-DS. Across all datasets, the visibility threshold $\tau$ is chosen from the range $[10^{-4}, 10^{-3}]$.

\section{Ablation Study Details}
\label{sec:abl_impl_details}

In this section, we describe the detailed implementation of the ablation study. All ablation experiments are conducted on top of the RTD model~\cite{wu20234dgs} and are evaluated on the D-NeRF and HyperNeRF datasets.
Unless otherwise specified, all ablation settings share the same hyperparameters: $\lambda_{SR}=1.0$, $\lambda_{MCR}=0.005$, $\tau=1\times10^{-3}$, and $m=20.0$. All other settings are kept identical across ablations for fair comparison.

For the KNN-based grouping, we follow the implementation of D3DGS~\cite{luiten2023dynamic}, which constructs groups by retrieving the $k$ nearest neighbors for each Gaussian. We set $k=20$ in all experiments. A key difference from D3DGS is that their method optimizes the parameters at timestep $t$ while fixing those at the previous timestep $t-1$, whereas our formulation jointly optimizes the parameters at two sampled timesteps, $t$ and $t+\Delta t$. We adopt this design because, in RTD, the Gaussian parameters at both timesteps are obtained by deforming the same canonical representation and are therefore implicitly coupled. The KNN search is implemented using Open3D.

For ARAP regularization, we implement both KNN-based and ray-based variants. In the KNN setting, we compute the cross-covariance of grouped Gaussian positions across two timesteps and recover the local rotation using singular value decomposition (SVD). The resulting rotation is then used in the ARAP loss term. For ray-based grouping, we use the same formulation based on cross-covariance between the positions at the two timesteps, but implement the computation in CUDA for efficiency. The resulting cross-covariance is likewise decomposed with SVD and used to compute the ARAP loss. Note that ARAP is evaluated independently and is not combined with MCR in the ablation study.
    
\section{Our Formulation of Welford's Algorithm}
\label{sec:welford_proof}

Gaussian Splatting processes Gaussians along each ray in ascending order of their distance from the camera. When computing the covariance of the means of Gaussians along a ray, a conventional two-pass algorithm needs to store all intermediate values in memory and run the rasterization pipeline multiple times. As a result, calculating the covariance at the ray level efficiently is a difficult task.

To address this challenge, we propose employing Welford’s algorithm to compute the covariance of the means of Gaussians along a ray, in conjunction with a method for calculating the partial derivatives necessary for the backward pass. Welford’s algorithm is particularly suitable as it enables covariance computation in a single pass. By applying this method, the covariance can be obtained by sequentially referencing each Gaussian exactly once.

In the following sections, we first present an overview of Welford’s algorithm and then derive the partial derivative with respect to the Gaussian mean. Notably, because the algorithm is based on a recurrence relation, it enables covariance computation without requiring multiple passes. When incorporated into training, it further supports both forward and backward passes with a limited memory footprint. 

\subsection{Derivation}
To compute the covariance of sequential data, Welford's algorithm~\cite{Welford1962Note} employs a recurrence relation.  
For the $n$-th Gaussian mean $\boldsymbol{\mu}_n$ along a ray, the mean up to step $n$ is defined as
\begin{equation}
\bar{\boldsymbol{\mu}}_n = \frac{1}{n} \sum_{i=1}^{n} \boldsymbol{\mu}_i .
\end{equation}

The covariance at step $n$ is defined as
\begin{equation}
C_n = \frac{1}{n} \sum_{i=1}^n \bigl(\boldsymbol{\mu}_i - \bar{\boldsymbol{\mu}}_n \bigr) 
\bigl(\boldsymbol{\mu}_i - \bar{\boldsymbol{\mu}}_n \bigr)^{\top}.
\end{equation}
Utilizing the recursive relationship for the mean, $\bar{\boldsymbol{\mu}}_{n}=\frac{n-1}{n}\bar{\boldsymbol{\mu}}_{n-1}+\frac{1}{n}\boldsymbol{\mu}_n$ the covariance $C_n$ can be expressed as
{\footnotesize \thinmuskip=2mu \medmuskip=2mu \thickmuskip=3mu 
\begin{align*}
C_n 
&= \frac{1}{n}\bigg[
\sum_{i=1}^{n-1} \bigl(\boldsymbol{\mu}_i - \bar{\boldsymbol{\mu}}_n \bigr) 
\bigl(\boldsymbol{\mu}_i - \bar{\boldsymbol{\mu}}_n \bigr)^{\top} + \bigl (\boldsymbol{\mu}_n - \bar{\boldsymbol{\mu}}_{n} \bigr) \bigl (\boldsymbol{\mu}_n - \bar{\boldsymbol{\mu}}_{n} \bigr)^{\top}
\bigg] \tag{\refstepcounter{equation}\theequation}
\\
&= \frac{1}{n}\bigg[
\sum_{i=1}^{n-1} (\boldsymbol{\mu}_i -\frac{n-1}{n}\bar{\boldsymbol{\mu}}_{n-1}-\frac{1}{n}\boldsymbol{\mu}_n)(\boldsymbol{\mu}_i - \frac{n-1}{n}\bar{\boldsymbol{\mu}}_{n-1}-\frac{1}{n}\boldsymbol{\mu}_n
) ^{\top} + \bigl (\boldsymbol{\mu}_n - \bar{\boldsymbol{\mu}}_{n} \bigr) \bigl (\boldsymbol{\mu}_n - \bar{\boldsymbol{\mu}}_{n} \bigr)^{\top}
\bigg] \tag{\refstepcounter{equation}\theequation}
\\
&= \frac{1}{n}\bigg[\sum_{i=1}^{n-1} \Bigl((\boldsymbol{\mu}_i -\bar{\boldsymbol{\mu}}_{n-1})-\frac{1}{n}(\boldsymbol{\mu}_n-\bar{\boldsymbol{\mu}}_{n-1}) \Bigr) \Bigl((\boldsymbol{\mu}_i - \bar{\boldsymbol{\mu}}_{n-1})-\frac{1}{n}(\boldsymbol{\mu}_n-\bar{\boldsymbol{\mu}}_{n-1}) \Bigr) ^{\top} \\
& ~\quad + \bigl (\boldsymbol{\mu}_n - \bar{\boldsymbol{\mu}}_{n} \bigr) \bigl (\boldsymbol{\mu}_n - \bar{\boldsymbol{\mu}}_{n} \bigr)^{\top}\bigg] \tag{\refstepcounter{equation}\theequation}
\\
&=\frac{1}{n}\bigg[\sum_{i=1}^{n-1} \Big \{
(\boldsymbol{\mu}_i - \bar{\boldsymbol{\mu}}_{n-1})(\boldsymbol{\mu}_i - \bar{\boldsymbol{\mu}}_{n-1}) ^{\top} 
- \frac{1}{n}(\boldsymbol{\mu}_i - \bar{\boldsymbol{\mu}}_{n-1})(\boldsymbol{\mu}_n - \bar{\boldsymbol{\mu}}_{n-1}) ^{\top}\\
& ~\quad- \frac{1}{n}(\boldsymbol{\mu}_n - \bar{\boldsymbol{\mu}}_{n-1}) \bigl (\boldsymbol{\mu}_i - \bar{\boldsymbol{\mu}}_{n-1})^{\top} 
+ \frac{1}{n^2}(\boldsymbol{\mu}_n - \bar{\boldsymbol{\mu}}_{n-1})(\boldsymbol{\mu}_n - \bar{\boldsymbol{\mu}}_{n-1}) ^{\top}
\Big\} \\
& ~\quad
 + \bigl (\boldsymbol{\mu}_n - \bar{\boldsymbol{\mu}}_{n} \bigr) \bigl (\boldsymbol{\mu}_n - \bar{\boldsymbol{\mu}}_{n} \bigr)^{\top}\bigg] \tag{\refstepcounter{equation}\theequation} 
\\
&= \frac{1}{n}\bigg[(n-1)C_{n-1}
- \frac{1}{n}\sum_{i=1}^{n-1} (\boldsymbol{\mu}_i - \bar{\boldsymbol{\mu}}_{n-1})(\boldsymbol{\mu}_n - \bar{\boldsymbol{\mu}}_{n-1})^\top
- \frac{1}{n}\sum_{i=1}^{n-1} (\boldsymbol{\mu}_n - \bar{\boldsymbol{\mu}}_{n-1})(\boldsymbol{\mu}_i - \bar{\boldsymbol{\mu}}_{n-1})^\top 
\\
& \quad + \frac{1}{n^2}\sum_{i=1}^{n-1} (\boldsymbol{\mu}_n - \bar{\boldsymbol{\mu}}_{n-1})(\boldsymbol{\mu}_n - \bar{\boldsymbol{\mu}}_{n-1})^\top 
+ \bigl (\boldsymbol{\mu}_n - \bar{\boldsymbol{\mu}}_{n} \bigr) \bigl (\boldsymbol{\mu}_n - \bar{\boldsymbol{\mu}}_{n} \bigr)^{\top}
\bigg]. \tag{\refstepcounter{equation}\theequation}
\intertext{{\normalsize Since} $\sum_{i=1}^{n-1} (\boldsymbol{\mu}_i - \bar{\boldsymbol{\mu}}_{n-1})=0$,}
C_n 
&= \frac{1}{n}\bigg[(n-1)C_{n-1}
- \frac{1}{n^2}\sum_{i=1}^{n-1} (\boldsymbol{\mu}_i - \bar{\boldsymbol{\mu}}_{n-1})(\boldsymbol{\mu}_n - \bar{\boldsymbol{\mu}}_{n-1})^\top 
+ \bigl (\boldsymbol{\mu}_n - \bar{\boldsymbol{\mu}}_{n} \bigr) \bigl (\boldsymbol{\mu}_n - \bar{\boldsymbol{\mu}}_{n} \bigr)^{\top}\bigg]. \tag{\refstepcounter{equation}\theequation}
\end{align*}
{\normalsize Using the identity}
$(\boldsymbol{\mu}_n - \bar{\boldsymbol{\mu}}_{n})(\boldsymbol{\mu}_n - \bar{\boldsymbol{\mu}}_{n})^\top=\bigl(\frac{n-1}{n}\bigr)^2(\boldsymbol{\mu}_n - \bar{\boldsymbol{\mu}}_{n-1})(\boldsymbol{\mu}_n - \bar{\boldsymbol{\mu}}_{n-1})^\top$,

\begin{equation}
C_n = \frac{1}{n}\Bigl[(n-1)C_{n-1} + \frac{n-1}{n}(\boldsymbol{\mu}_n - \bar{\boldsymbol{\mu}}_{n-1})(\boldsymbol{\mu}_n - \bar{\boldsymbol{\mu}}_{n-1})^\top\Bigr].
\end{equation}
}
Therefore, recurrence relation of covariance is
{\footnotesize
\begin{equation}
C_n = \frac{n-1}{n}C_{n-1} + \frac{1}{n}(\boldsymbol{\mu}_n - \bar{\boldsymbol{\mu}}_{n-1})
(\boldsymbol{\mu}_n - \bar{\boldsymbol{\mu}}_n)^{\top}.
\end{equation}
}

\subsection{Covariance Gradient w.r.t. a Mean}
If the number of Gaussians along a ray is $N$, the covariance is given by  
{\footnotesize
\begin{equation}
C_N = \frac{N-1}{N} C_{N-1}
      + \frac{1}{N}(\boldsymbol{\mu}_N - \bar{\boldsymbol{\mu}}_{N-1})
        (\boldsymbol{\mu}_N - \bar{\boldsymbol{\mu}}_N)^{\top}.
\end{equation}
}
Starting from the recursive definition
{\footnotesize
\begin{equation}
C_i = \frac{i-1}{i}C_{i-1} + \frac{1}{i}P_i, 
\quad P_i := \bigl(\boldsymbol{\mu}_i - \bar{\boldsymbol{\mu}}_{\,i-1}\bigr)
       \bigl(\boldsymbol{\mu}_i - \bar{\boldsymbol{\mu}}_i \bigr)^{\top},
\quad i=1,\dots,N.
\end{equation}
}
By induction, the $C_N$ can be written as
{\footnotesize
\begin{equation}
C_N = \sum_{i=1}^N 
\left(\frac{1}{i}\prod_{j=i+1}^N \frac{j-1}{j}\right)P_i
= \frac{1}{N}\sum_{i=1}^N P_i.
\end{equation}
}
The derivative of $C_N$ with respect to the $k$-th component of the $n$-th Gaussian mean vector is
{\footnotesize
\begin{equation}
\frac{\partial C_N}{\partial \mu_{n,k}}
= \frac{1}{N}\sum_{i=1}^N \frac{\partial P_i}{\partial \mu_{n,k}}
= \frac{1}{N}\sum_{i=n}^N \frac{\partial P_i}{\partial \mu_{n,k}}
\end{equation}
}
where $\mu_{n,k}$ is the $k$-th component of the vector $\boldsymbol{\mu}_n$. This holds because $P_i$ does not depend on $\mu_{n,k}$ when $i<n$.

\vspace{1em}
\noindent\textbf{Case 1: $i=n$.} \\
Using $\dfrac{\partial \bar{\boldsymbol{\mu}}_n}{\partial \mu_{n,k}} = \dfrac{1}{n}\mathbf{e}_k$, where $\mathbf{e}_k$ denotes the $k$-th standard basis vector in $\mathbb{R}^3$, we obtain
{\footnotesize
\begin{equation}
\frac{\partial P_n}{\partial \mu_{n,k}} = \frac{n-1}{n}\bigg[\mathbf{e}_k\,(\boldsymbol{\mu}_n - \bar{\boldsymbol{\mu}}_{n-1})^{\top} + (\boldsymbol{\mu}_n - \bar{\boldsymbol{\mu}}_{\,n-1})\,\mathbf{e}_k^{\top}\bigg].
\end{equation}
}

\vspace{1em}
\noindent\textbf{Case 2: $i>n$.} \\
Here both $\bar{\boldsymbol{\mu}}_{i-1}$ and $\bar{\boldsymbol{\mu}}_{n}$ depend on $\boldsymbol{\mu}_n$.
Since 
$\dfrac{\partial \bar{\boldsymbol{\mu}}_{i-1}}{\partial \mu_{n,k}} 
= \dfrac{1}{i-1}\mathbf{e}_k$
and 
$\dfrac{\partial \bar{\boldsymbol{\mu}}_{i}}{\partial \mu_{n,k}} 
= \dfrac{1}{i}\mathbf{e}_k$,
we obtain
{\footnotesize
\begin{equation}
\frac{\partial P_i}{\partial \boldsymbol{\mu}_{n,k}}
= -\frac{1}{i-1}\,\mathbf{e}_k\,(\boldsymbol{\mu}_i - \bar{\boldsymbol{\mu}}_i)^{\top}
-\frac{1}{i}\,(\boldsymbol{\mu}_i - \bar{\boldsymbol{\mu}}_{\,i-1})\,\mathbf{e}_k^{\top}.
\end{equation}
}

\vspace{1em}
\noindent\textbf{Final Result.} \\
Combining both cases, we obtain
{\footnotesize
\begin{align*}
\frac{\partial C_N}{\partial \mu_{n,k}}
&=\frac{1}{N}\Bigg[
\mathbf{e}_k\,(\boldsymbol{\mu}_n - \bar{\boldsymbol{\mu}}_n)^{\top}
+ \frac{n-1}{n}\,(\boldsymbol{\mu}_n - \bar{\boldsymbol{\mu}}_{\,n-1})\,\mathbf{e}_k^{\top} \\
&- \sum_{i=n+1}^{N}
\left(
\frac{1}{i-1}\,\mathbf{e}_k\,(\boldsymbol{\mu}_i - \bar{\boldsymbol{\mu}}_i)^{\top}
+ \frac{1}{i}\,(\boldsymbol{\mu}_i - \bar{\boldsymbol{\mu}}_{\,i-1})\,\mathbf{e}_k^{\top}
\right)
\Bigg].  \tag{\refstepcounter{equation}\theequation}
\end{align*}}
For convenience, define, 
{\footnotesize
\begin{equation}
{R}_k(i)
= \frac{1}{i-1}\,\mathbf{e}_k\,(\boldsymbol{\mu}_i-\bar{\boldsymbol{\mu}}_{\,i})^{\top}
\;+\; \frac{1}{i}\,(\boldsymbol{\mu}_i-\bar{\boldsymbol{\mu}}_{\,i-1})\,\mathbf{e}_k^{\top},
\qquad {R}(1)=\mathbf{0}.
\end{equation}
}
The derivative can then be compactly expressed as
{\footnotesize
\begin{equation}
\frac{\partial C_N}{\partial \mu_{n,k}}
\;=\;
\frac{1}{N}\left[
  (n-1)\,{R}_k(n)
  -
  \sum_{i=n+1}^{N}{R}_k(i)
\right].
\end{equation}
}

\vspace{5mm}

\section{Additional Per-Scene Evaluation Results}
In this section, we report per-scene evaluation results of our approach on D-NeRF (\cref{tab:performance_comparison_supp1}), HyperNeRF (\cref{tab:performance_comparison_supp2}) and NeRF-DS datasets (\cref{tab:performance_comparison_supp3}). In addition, a video demonstration of our approach is provided in our supplementary material.

\label{sec:additional_vis}

\input{tables/scene_breakdown_dnerf}
\input{tables/scene_breakdown_hypernerf}
\input{tables/scene_breakdown_nerfds}

%% file: tables/scene_breakdown_dnerf.tex
\begin{table*}[h]
\centering
\caption{Quantitative comparison of our method on the D-NeRF dataset. We report PSNR, SSIM and LPIPS (VGG). {\small\textdagger}: Evaluation on validation set.} %
\label{tab:performance_comparison_supp1}
\resizebox{1.0\linewidth}{!} {%
\begin{tabular}{l c ccc c ccc c ccc c ccc}
\toprule
\multirow{3}{*}{\centering Model} & ~~~ & \multicolumn{3}{c}{\tworow{Bouncing}{Balls} } & ~~~ & \multicolumn{3}{c}{\tworow{Hell}{Warrior}} & ~~~ & \multicolumn{3}{c}{Hook} & ~~~ & \multicolumn{3}{c}{\tworow{Jumping}{Jacks}} \\
\cmidrule(){3-5} \cmidrule(){7-9} \cmidrule(){11-13} \cmidrule(){15-17}
& & PSNR$\uparrow$ \, & \!SSIM$\uparrow$\! \, & LPIPS$_{A}\!\downarrow$ \, & ~~~ & PSNR$\uparrow$ \, & \!SSIM$\uparrow$\! \, & LPIPS$_{A}\!\downarrow$ \, & ~~~ & PSNR$\uparrow$ \, & \!SSIM$\uparrow$\! \, & LPIPS$_{A}\!\downarrow$ & & ~~~ PSNR$\uparrow$ \, & \!SSIM$\uparrow$\! \, & LPIPS$_{A}\!\downarrow$ \\
\midrule
NeRF~\cite{mildenhall2020nerf} & & 20.26 & 0.91  & 0.20  &  & 13.52 & 0.81  & 0.25  &  & 16.65 & 0.84  & 0.19  &  & 18.28 & 0.88  & 0.23  \\
D-NeRF~\cite{pumarola2020dnerf} & & 38.93 & 0.98  & 0.03  &  & 25.02 & 0.95  & 0.06  &  & 29.25 & 0.96  & 0.11  &  & 32.80 & 0.98  & 0.03  \\
TiNeuVox~\cite{fang2022TiNeuVox} & & 40.73 & 0.99  & 0.04  &  & 28.17 & 0.97  & 0.07  &  & 31.45 & 0.97  & 0.05  &  & 34.23 & 0.98  & 0.03  \\
3D-GS~\cite{kerbl20233d} & & 23.20 & 0.959 & 0.060 &  & 29.89 & 0.916 & 0.106 &  & 21.71 & 0.888 & 0.103 &  & 20.64 & 0.930 & 0.083 \\
D3DGS~\cite{yang2023deformable} & & 41.01 & \tul{0.995} & 0.009 &  & 41.54 & \tul{0.987} & 0.023 &  & 37.42 & 0.987 & 0.014 &  & 37.72 & 0.990 & 0.013 \\
GaGS~\cite{lu20243d} & & 
\tbf{43.59} & \tbf{0.996} & \tbf{0.006} & & 32.27 & 0.984 & \tul{0.016} & & 36.96 & \tbf{0.992} & \tbf{0.008} & & 37.96 & \tbf{0.993} & 0.009  \\

\midrule
Ex4DGS~\cite{lee2024fully} & & 34.51 & 0.988 & 0.029 &  & 23.44 & 0.946 & 0.070 &  & 27.32 & 0.948 & 0.054 &  & 30.30 & 0.969 & 0.043 \\

RTD~\cite{wu20234dgs} & & 40.84 & \tul{0.995} & 0.014 &  & 28.68 & 0.973 & 0.038 &  & 33.03 & 0.977 & 0.028 &  & 35.46 & 0.986 & 0.020 \\

MoDec-GS~\cite{kwak2025modecgs} & & 35.34 & 0.986 & 0.037 &  & 21.04 & 0.934 & 0.082 &  & 24.63 & 0.931 & 0.069 &  & 32.30 & 0.976 & 0.036 \\

Grid4D~\cite{xu2024grid4d} & & 
42.46 & \tbf{0.996} & 0.008 &  & \tul{42.87} & \tbf{0.991} & \tul{0.016} &  & \tul{39.00} & \tul{0.990} & \tul{0.009} &  & \tul{39.27} & \tul{0.992} & \tul{0.008} \\\midrule

Ex4DGS$+$\tbf{Ours} & & 35.60 & 0.990 & 0.026 &  & 25.84 & 0.955 & 0.069 &  & 29.38 & 0.961 & 0.042 &  & 31.66 & 0.972 & 0.041 \\

RTD$+$\tbf{Ours} & & 41.57 & \tul{0.995} & 0.013 &  & 29.82 & 0.977 & 0.033 &  & 33.93 & 0.980 & 0.025 &  & 36.40 & 0.988 & 0.018 \\

MoDec-GS$+$\tbf{Ours} & & 38.25 & 0.991 & 0.022 &  & 24.36 & 0.958 & 0.054 &  & 27.86 & 0.953 & 0.048 &  & 32.36 & 0.978 & 0.033 \\

Grid4D$+$\tbf{Ours} & & \tul{42.68} & \tbf{0.996} & \tul{0.007} &  & \tbf{43.08} & \tbf{0.991} & \tbf{0.015} &  & \tbf{39.01} & \tul{0.990} & \tul{0.009} &  & \tbf{39.54} & \tbf{0.993} & \tbf{0.007} \\

\midrule \midrule
\multirow{3}{*}{\centering Model} & ~~~ & \multicolumn{3}{c}{Lego } & ~~~ & \multicolumn{3}{c}{Mutant} & ~~~ & \multicolumn{3}{c}{Standup} & ~~~ & \multicolumn{3}{c}{Trex} \\
\cmidrule(){3-5} \cmidrule(){7-9} \cmidrule(){11-13} \cmidrule(){15-17}
& & PSNR$\uparrow$ \, & \!SSIM$\uparrow$\! \, & LPIPS$_{A}\!\downarrow$ \, & ~~~ & PSNR$\uparrow$ \, & \!SSIM$\uparrow$\! \, & LPIPS$_{A}\!\downarrow$ \, & ~~~ & PSNR$\uparrow$ \, & \!SSIM$\uparrow$\! \, & LPIPS$_{A}\!\downarrow$ & & ~~~ PSNR$\uparrow$ \, & \!SSIM$\uparrow$\! \, & LPIPS$_{A}\!\downarrow$ \\
\midrule
NeRF~\cite{mildenhall2020nerf} & & 20.30 & 0.79 & 0.23 & & 20.31 & 0.91 & 0.09 & & 18.19 & 0.89 & 0.14 & & 24.49 & 0.93 & 0.13  \\
D-NeRF~\cite{pumarola2020dnerf} & & 21.64 & 0.83 & 0.16 & & 31.29 & 0.97 & 0.02 & & 32.79 & 0.98 & 0.02 & & 31.75 & 0.97 & 0.03  \\
TiNeuVox~\cite{fang2022TiNeuVox} & & 25.02 & 0.92 & 0.07 & & 33.61 & 0.98 & 0.03 & & 35.43 & 0.99 & 0.02 & & 32.70 & 0.98 & 0.03  \\
3D-GS~\cite{kerbl20233d} & & 22.10 & 0.938 & 0.061 & & 24.53 & 0.934 & 0.058 & & 21.91 & 0.930 & 0.079 & & 21.93 & 0.954 & 0.049  \\
D3DGS~\cite{yang2023deformable} & & 33.07\textsuperscript{\textdagger} & 0.979\textsuperscript{\textdagger} & 0.018\textsuperscript{\textdagger} & & 42.63 & 0.995 & 0.005 & & 44.62 & \tul{0.995} & 0.006 & & 38.10 & \tul{0.993} & 0.010  \\
GaGS~\cite{lu20243d} & & \tbf{25.44} & \tbf{0.947} & \tbf{0.033}  & & 41.43 & \tbf{0.997} & \tbf{0.003} & & 42.21 & \tbf{0.997} & \tbf{0.003} & & 39.03 & \tbf{0.995} & \tbf{0.005}  \\
\midrule
Ex4DGS~\cite{lee2024fully} & & 24.96 & 0.941 & 0.048 &  & 33.04 & 0.979 & 0.025 &  & 32.22 & 0.979 & 0.028 &  & 28.65 & 0.969 & 0.035 \\

RTD~\cite{wu20234dgs} & & 25.02 & 0.937 & 0.059 &  & 37.42 & 0.987 & 0.019 &  & 37.97 & 0.990 & 0.014 &  & 34.13 & 0.984 & 0.023 \\

MoDec-GS~\cite{kwak2025modecgs} & & 20.39 & 0.862 & 0.130 &  & 32.92 & 0.976 & 0.032 &  & 30.36 & 0.971 & 0.040 &  & 28.69 & 0.966 & 0.042 \\

Grid4D~\cite{xu2024grid4d} & & 24.84 & 0.942 & \tul{0.045} &  & \tul{43.99} & \tul{0.996} & \tul{0.004} &  & \tul{46.37} & \tbf{0.997} & \tul{0.004} &  & \tul{40.01} & \tbf{0.995} & \tul{0.008} \\
\midrule

Ex4DGS$+$\tbf{Ours} & & \tul{25.17} & \tul{0.943} & 0.050 &  & 32.95 & 0.978 & 0.026 &  & 32.22 & 0.976 & 0.033 &  & 29.62 & 0.973 & 0.032 \\
RTD$+$\tbf{Ours} & & 25.08 & 0.938 & 0.057 &  & 38.49 & 0.990 & 0.015 &  & 39.22 & 0.991 & 0.012 &  & 35.81 & 0.988 & 0.020 \\
MoDec-GS$+$\tbf{Ours} & & 23.76 & 0.922 & 0.070 &  & 33.94 & 0.979 & 0.027 &  & 33.44 & 0.982 & 0.024 &  & 31.56 & 0.977 & 0.033 \\
Grid4D$+$\tbf{Ours} & & 24.90 & 0.942 & \tul{0.045} &  & \tbf{44.02} & \tul{0.996} & \tul{0.004} &  & \tbf{46.61} & \tbf{0.997} & \tul{0.004} &  & \tbf{40.48} & \tbf{0.995} & \tul{0.008} \\

\bottomrule
\end{tabular}%
} %
\end{table*}

%% file: tables/scene_breakdown_hypernerf.tex
\begin{table*}[h]
\centering
\caption{Quantitative comparison of our method on the HyperNeRF dataset. We report PSNR, MS-SSIM and LPIPS (Alex).} %
\label{tab:performance_comparison_supp2}
\resizebox{1.0\linewidth}{!} {%
\begin{tabular}{l c ccc c ccc c ccc c ccc}
\toprule
\multirow{3}{*}{\centering Model} & ~~~ & \multicolumn{3}{c}{3D Printer} & ~~~ & \multicolumn{3}{c}{Broom} & ~~~ & \multicolumn{3}{c}{Chicken} & ~~~ & \multicolumn{3}{c}{Peel Banana} \\
\cmidrule(){3-5} \cmidrule(){7-9} \cmidrule(){11-13} \cmidrule(){15-17}
& & PSNR$\uparrow$\, \, & \!\!MS-SSIM$\uparrow$\!\! \, & LPIPS$_{A}\!\downarrow$ \, & ~~~ & PSNR$\uparrow$\, \, & \!\!MS-SSIM$\uparrow$\!\! \, & LPIPS$_{A}\!\downarrow$ \, & ~~~ & PSNR$\uparrow$\, \, & \!\!MS-SSIM$\uparrow$\!\! \, & LPIPS$_{A}\!\downarrow$ & & ~~~ PSNR$\uparrow$\, \, & \!\!MS-SSIM$\uparrow$\!\! \, & LPIPS$_{A}\!\downarrow$ \\
\midrule
NeRF~\cite{mildenhall2020nerf} & & 20.7 & 0.780 & 0.357 & & 19.9 & 0.653 & 0.692 & & 19.9 & 0.777 & 0.325 & & 20.0 & 0.769 & 0.413  \\
TiNeuVox~\cite{fang2022TiNeuVox} & & \tbf{22.8} & \tbf{0.841} & \NA & & 21.5 & 0.686 & \NA & & 28.3 & \tul{0.947} & \NA & & 24.4 & 0.873 & \NA  \\
HyperNeRF~\cite{park2021hypernerf} & & 20.0 & 0.821 & \NA & & 19.3 & 0.591 & \NA & & 26.9 & \tbf{0.948} & \NA & & 23.3 & 0.896 & \NA  \\
3D-GS~\cite{kerbl20233d} & & 19.26 & 0.669 & \NA & & 19.74 & 0.495 & \NA & & 22.51 & 0.795 & \NA & & 19.54 & 0.669 & \NA  \\
D3DGS~\cite{yang2023deformable} & & 20.38 & \NA & \NA & & 20.48 & \NA & \NA & & 22.64 & \NA & \NA & & 26.10 & \NA & \NA  \\
GaGS~\cite{lu20243d} & & 22.04 & 0.810 & \NA & & 20.90 & 0.524 & \NA & & 28.53 & 0.933 & \NA & & 25.58 & 0.907 & \NA  \\
\midrule
Ex4DGS~\cite{lee2024fully} & & 22.36 & 0.816 & 0.201 &  & 20.92 & 0.642 & 0.321 &  & 28.55 & 0.939 & 0.109 &  & 26.59 & 0.919 & 0.164  \\
RTD~\cite{wu20234dgs} & & 22.08 & 0.807 & 0.269 &  & \tul{22.01} & 0.689 & 0.548 &  & 28.56 & 0.929 & 0.178 &  & 28.02 & 0.941 & 0.135  \\
MoDec-GS~\cite{kwak2025modecgs} & & 21.92 & 0.806 & 0.188 &  & 21.13 & 0.678 & \tul{0.289} &  & 28.61 & 0.934 & \tul{0.107} &  & 28.24 & \tul{0.947} & \tul{0.120}  \\
Grid4D~\cite{xu2024grid4d} & & 22.29 & 0.824 & \tul{0.177} &  & 21.86 & 0.416 & 0.328 &  & \tul{29.30} & 0.943 & 0.123 &  & \tul{28.57} & \tul{0.947} & \tbf{0.111}  \\
\midrule
Ex4DGS$+$\tbf{Ours} & & \tul{22.41} & 0.818 & 0.230 &  & 21.50 & 0.665 & 0.449 &  & 28.63 & 0.940 & 0.110 &  & 26.90 & 0.922 & 0.143  \\
RTD$+$\tbf{Ours} & & 22.12 & 0.811 & 0.247 &  & \tbf{22.08} & \tul{0.694} & 0.552 &  & 28.68 & 0.932 & 0.167 &  & 28.34 & 0.944 & 0.129  \\
MoDec-GS$+$\tbf{Ours} & & 22.07 & 0.815 & 0.196 &  & 21.24 & 0.688 & \tbf{0.274} &  & 28.74 & 0.936 & \tbf{0.103} &  & 28.31 & \tbf{0.948} & \tul{0.120}  \\
Grid4D$+$\tbf{Ours} & & 22.34 & \tul{0.825} & \tbf{0.176} &  & 21.94 & \tbf{0.709} & 0.385 &  & \tbf{29.33} & 0.944 & 0.119 &  & \tbf{28.61} & \tbf{0.948} & \tbf{0.111}  \\
\bottomrule
\end{tabular}%
} %
\end{table*}

%% file: tables/scene_breakdown_nerfds.tex
\begin{table*}[h]
\centering
\caption{Quantitative comparison of our method on the NeRF-DS dataset. We report PSNR, SSIM, MS-SSIM and LPIPS (VGG). Despite the optimization challenges posed by imperfect camera poses and the train/test split in some scenes, our method provides clear overall improvements across baselines.} %
\label{tab:performance_comparison_supp3}
\resizebox{1.0\linewidth}{!} {%
\begin{tabular}{l c ccc c ccc c ccc c ccc}
\toprule
\multirow{3}{*}{\centering Model} & ~~~ & \multicolumn{3}{c}{As } & ~~~ & \multicolumn{3}{c}{Basin} & ~~~ & \multicolumn{3}{c}{Bell} & ~~~ & \multicolumn{3}{c}{Cup} \\
\cmidrule(){3-5} \cmidrule(){7-9} \cmidrule(){11-13} \cmidrule(){15-17}
& & PSNR$\uparrow$\,\, & SSIM/MS$\uparrow$ & \! \!LPIPS$_{V}\!\downarrow$ & ~~~ & PSNR$\uparrow$\,\, & SSIM/MS$\uparrow$ & \! \!LPIPS$_{V}\!\downarrow$ & ~~~ & PSNR$\uparrow$\,\, & SSIM/MS$\uparrow$ & \! \!LPIPS$_{V}\!\downarrow$ & ~~~ &  PSNR$\uparrow$\,\, & SSIM/MS$\uparrow$ & \! \!LPIPS$_{V}\!\downarrow$ \\
\midrule
TiNeuVox~\cite{fang2022TiNeuVox}   & & 21.26 & 0.829/~~~-~~~ & 0.397 & & \tbf{20.66} & \tbf{0.815}/~~~-~~~ & 0.269 & & 23.08 & 0.824/~~~-~~~ & 0.257 & & 19.71 & 0.811/~~~-~~~ & 0.364  \\
HyperNeRF~\cite{park2021hypernerf} & & 25.58 & ~~~-~~~/0.895 & \tul{0.178} & & \tul{20.41} & ~~~-~~~/0.820 & 0.191 & & 23.06 & ~~~-~~~/0.810 & 0.205 & & 24.59 & ~~~-~~~/0.877 & 0.165  \\
NeRF-DS~\cite{yan2023nerfds}       & & 25.13 & ~~~-~~~/0.878 & \tbf{0.174} & & 19.96 & ~~~-~~~/0.817 & \tbf{0.186} & & 23.19 & ~~~-~~~/0.821 & 0.187 & & \tbf{24.91} & ~~~-~~~/0.874 & 0.174  \\
3D-GS~\cite{kerbl20233d}           & & 22.69 & 0.802/~~~-~~~ & 0.299 & & 18.42 & 0.717/~~~-~~~ & 0.315 & & 21.01 & 0.789/~~~-~~~ & 0.250 & & 23.16 & 0.830/~~~-~~~ & 0.255  \\
D3DGS~\cite{yang2023deformable}    & & \tul{26.08} & \tul{0.883}/~~~-~~~ & 0.183 & & 19.61 & \tul{0.789}/~~~-~~~ & \tul{0.187} & & \tbf{25.42} & \tbf{0.848}/~~~-~~~ & \tbf{0.157} & & \tul{24.76} & \tbf{0.888}/~~~-~~~ & \tbf{0.154}  \\\midrule
Ex4DGS~\cite{lee2024fully}         & & 22.93 & 0.826/0.794 & 0.244 &  & 18.73 & 0.744/0.756 & 0.253 &  & 22.69 & 0.783/0.836 & 0.221 &  & 23.06 & 0.848/0.842 & 0.204 \\
RTD~\cite{wu20234dgs}              & & 22.62 & 0.831/0.809 & 0.251 &  & 19.07 & 0.759/0.800 & 0.256 &  & 21.28 & 0.788/0.812 & 0.235 &  & 23.93 & 0.873/0.879 & 0.178 \\
MoDec-GS~\cite{kwak2025modecgs}    & & 25.95 & \tul{0.883}/\tul{0.898} & 0.183 &  & 19.66 & 0.783/0.833 & 0.199 &  & 21.76 & 0.737/0.778 & 0.347 &  & 23.93 & 0.883/0.909 & \tul{0.163} \\
Grid4D~\cite{xu2024grid4d}         & & 25.77 & 0.862/0.881 & 0.201 &  & 19.34 & 0.776/0.828 & 0.200 &  & 24.98 & \tul{0.838}/\tbf{0.916} & 0.170 &  & 24.06 & 0.876/0.909  & 0.169 \\\midrule
Ex4DGS$+$\tbf{Ours}                & & 23.19 & 0.830/0.799 & 0.243 &  & 18.86 & 0.750/0.768 & 0.248 &  & 22.79 & 0.787/0.838 & 0.218 &  & 23.02 & 0.849/0.842 & 0.203  \\
RTD$+$\tbf{Ours}                   & & 24.48 & 0.857/0.855 & 0.221 &  & 18.21 & 0.754/0.799 & 0.229 &  & 21.99 & 0.794/0.829 & 0.221 &  & 23.78 & 0.866/0.870 & 0.183  \\
MoDec-GS$+$\tbf{Ours}              & & \tbf{26.18} & \tbf{0.884}/\tbf{0.902} & 0.181 &  & 19.61 & 0.782/\tul{0.835} & 0.198 &  & 24.12 & 0.830/\tul{0.899} &\tul{0.165} &  & 24.36 & 0.880/\tul{0.910} & 0.164  \\
Grid4D$+$\tbf{Ours}                & & 25.95 & 0.870/0.887 & 0.188 &  & 19.52 & 0.783/\tbf{0.842} & 0.194 &  & \tul{25.18} & \tul{0.838}/\tbf{0.916} & 0.172 &  & 24.45 & \tul{0.885}/\tbf{0.919} & \tbf{0.154}  \\
\midrule \midrule
\multirow{3}{*}{\centering Model} & ~~~ & \multicolumn{3}{c}{Plate} & ~~~ & \multicolumn{3}{c}{Press} & ~~~ & \multicolumn{3}{c}{Sieve} & ~~~ & \multicolumn{3}{c}{} \\
\cmidrule(){3-5} \cmidrule(){7-9} \cmidrule(){11-13} 
& & PSNR$\uparrow$\,\, & SSIM/MS$\uparrow$ & \!LPIPS$_{V}\!\downarrow$ & ~~~ & PSNR$\uparrow$\,\, & SSIM/MS$\uparrow$ & \!LPIPS$_{V}\!\downarrow$ & ~~~ & PSNR$\uparrow$\,\, & SSIM/MS$\uparrow$ & \!LPIPS$_{V}\!\downarrow$ & & ~~~  &  & \\
\cmidrule(){0-12}
TiNeuVox~\cite{fang2022TiNeuVox}    & & \tbf{20.58} & 0.803/~~~-~~~ & 0.332 & & 24.47 & \tbf{0.861}/~~~-~~~ & 0.300 & & 21.49 & 0.827/~~~-~~~ & 0.318 & &  &  &   \\
HyperNeRF~\cite{park2021hypernerf}  & & 18.93 & ~~~-~~~/0.771 & 0.294 & & \tbf{26.15} & ~~~-~~~/\tbf{0.890} & \tul{0.196} & & 25.43 & ~~~-~~~/0.880 & 0.165 & &  &  &   \\
NeRF-DS~\cite{yan2023nerfds}        & & \tul{20.54} & ~~~-~~~/0.804 & \tbf{0.200} & & 25.72 & ~~~-~~~/0.862 & 0.205 & & \tul{25.78} & ~~~-~~~/0.890 & \tbf{0.147} & &  &  &   \\
3D-GS~\cite{kerbl20233d}            & & 16.14 & 0.697/~~~-~~~ & 0.409 & & 22.89 & 0.816/~~~-~~~ & 0.290 & & 21.71 & 0.830/~~~-~~~ & 0.255 & &  &  &   \\
D3DGS~\cite{yang2023deformable}     & & 18.82 & 0.740/~~~-~~~ & 0.355 & & 25.41 & \tbf{0.861}/~~~-~~~ & \tbf{0.192} & & 25.14 & \tul{0.867}/~~~-~~~ & 0.150 & &  &  &   \\\cmidrule(){0-12}
Ex4DGS~\cite{lee2024fully}          & & 17.30 & 0.722/0.663 & 0.358 &  & 23.77 & 0.810/0.803 & 0.263 &  & 23.71 & 0.820/0.828 & 0.211 & &  &  &   \\
RTD~\cite{wu20234dgs}               & & 18.14 & 0.755/0.720 & 0.303 &  & 24.14 & 0.828/0.837 & 0.239 &  & 24.16 & 0.852/0.892 & 0.181 & &  &  &   \\
MoDec-GS~\cite{kwak2025modecgs}     & & 20.15 & \tul{0.808}/\tul{0.829} & 0.229 &  & 25.34 & 0.856/0.877 & 0.208 &  & 23.70 & 0.806/0.836 & 0.275 & &  &  &   \\
Grid4D~\cite{xu2024grid4d}          & & 19.28 & 0.774/0.785 & 0.267 &  & 25.18 & 0.836/0.877 & 0.210 &  & 25.01 & 0.860/0.909 & 0.155 & &  &  & \\\cmidrule(){0-12}
Ex4DGS$+$\tbf{Ours}                 & & 17.84 & 0.730/0.670 & 0.351 &  & 23.95 & 0.813/0.812 & 0.260 &  & 23.48 & 0.825/0.830 & 0.211 & &  &  &   \\
RTD$+$\tbf{Ours}                    & & 17.19 & 0.740/0.683 & 0.324 &  & 24.19 & 0.826/0.851 & 0.229 &  & 24.52 & 0.853/0.892 & 0.178 & &  &  &   \\
MoDec-GS$+$\tbf{Ours}               & & 20.48 & \tbf{0.817}/\tbf{0.844} & \tul{0.218} &  & \tul{25.75} & \tul{0.859}/0.884 & 0.206 &  & \tbf{25.83} & \tbf{0.880}/\tbf{0.923} & 0.158 & &  &  &   \\
Grid4D$+$\tbf{Ours}                 & & 19.77 & 0.788/0.805 & 0.258 &  & 25.68 & 0.849/\tul{0.886} & 0.203 &  & 25.36 & \tul{0.867}/\tul{0.918} & \tul{0.149} & &  &  &  \\
\bottomrule
\end{tabular}%
} %
\end{table*}

%% file: main.bib
@String(PAMI  = {IEEE Trans. Pattern Anal. Mach. Intell.})

@String(CVPR  = {IEEE Conf. Comput. Vis. Pattern Recog.})

@String(ICCV  = {Int. Conf. Comput. Vis.})

@String(ECCV  = {Eur. Conf. Comput. Vis.})

@String(NeurIPS = {Adv. Neural Inform. Process. Syst.})

@String(ICML  = {Int. Conf. Mach. Learn.})

@String(ICLR  = {Int. Conf. Learn. Represent.})

@String(CVPRW = {IEEE Conf. Comput. Vis. Pattern Recog. Worksh.})

@String(AAAI  = {AAAI})

@String(TOG   = {ACM Trans. Graph.})

@String(TVCG  = {IEEE Trans. Vis. Comput. Graph.})

@String(TDV   = {Int. Conf. 3D. Vis.})

@String(VIS   = {IEEE Vis.})

@String(PAMI  = {IEEE TPAMI})

@String(CVPR  = {CVPR})

@String(ICCV  = {ICCV})

@String(ECCV  = {ECCV})

@String(NeurIPS = {NeurIPS})

@String(ICML  = {ICML})

@String(ICLR  = {ICLR})

@String(CVPRW = {CVPRW})

@String(TOG   = {ACM TOG})

@String(TVCG  = {IEEE TVCG})

@String(TDV   = {3DV})

@String(SGP   = {SGP})

@String(VIS   = {VIS})

@String(SIGGRAPH    = {SIGGRAPH })

@inproceedings{lee2024geometry,
    title={Geometry-Aware Projective Mapping for Unbounded Neural Radiance Fields},
    author={Lee, Junoh and Jung, Hyunjun and Park, Jin-Hwi and Bae, Inhwan and Jeon, Hae-Gon},
    booktitle=ICLR,
    year={2024}
}

@inproceedings{mildenhall2020nerf,
    title={Nerf: Representing scenes as neural radiance fields for view synthesis},
    author={Mildenhall, B and Srinivasan, PP and Tancik, M and Barron, JT and Ramamoorthi, R and Ng, R},
    booktitle=ECCV,
    year={2020}
}

@article{kerbl20233d,
    title={3d gaussian splatting for real-time radiance field rendering},
    author={Kerbl, Bernhard and Kopanas, Georgios and Leimk{\"u}hler, Thomas and Drettakis, George},
    journal=TOG,
    year={2023}
}

@article{song2023nerfplayer,
    title={{NeRFPlayer}: A streamable dynamic scene representation with decomposed neural radiance fields},
    author={Song, Liangchen and Chen, Anpei and Li, Zhong and Chen, Zhang and Chen, Lele and Yuan, Junsong and Xu, Yi and Geiger, Andreas},
    journal=TVCG,
    year={2023}
}

@inproceedings{cao2023hexplane,
    title={Hexplane: A fast representation for dynamic scenes},
    author={Cao, Ang and Johnson, Justin},
    booktitle=CVPR,
    year={2023}
}

@inproceedings{fridovich2023k,
    title={K-{P}lanes: Explicit radiance fields in space, time, and appearance},
    author={Fridovich-Keil, Sara and Meanti, Giacomo and Warburg, Frederik Rahb{\ae}k and Recht, Benjamin and Kanazawa, Angjoo},
    booktitle=CVPR,
    year={2023}
}

@inproceedings{li2022neural,
    title={Neural 3d video synthesis from multi-view video},
    author={Li, Tianye and Slavcheva, Mira and Zollhoefer, Michael and Green, Simon and Lassner, Christoph and Kim, Changil and Schmidt, Tanner and Lovegrove, Steven and Goesele, Michael and Newcombe, Richard and others},
    booktitle=CVPR,
    year={2022}
}

@inproceedings{park2021nerfies,
    title={Nerfies: Deformable neural radiance fields},
    author={Park, Keunhong and Sinha, Utkarsh and Barron, Jonathan T and Bouaziz, Sofien and Goldman, Dan B and Seitz, Steven M and Martin-Brualla, Ricardo},
    booktitle=CVPR,
    year={2021}
}

@article{park2021hypernerf,
    title={HyperNeRF: a higher-dimensional representation for topologically varying neural radiance fields},
    author={Park, Keunhong and Sinha, Utkarsh and Hedman, Peter and Barron, Jonathan T and Bouaziz, Sofien and Goldman, Dan B and Martin-Brualla, Ricardo and Seitz, Steven M},
    journal=TOG,
    year={2021}
}

@inproceedings{luiten2023dynamic,
    title={Dynamic 3D Gaussians: Tracking by Persistent Dynamic View Synthesis},
    author={Luiten, Jonathon and Kopanas, Georgios and Leibe, Bastian and Ramanan, Deva},
    booktitle=TDV,
    year={2024}
}

@inproceedings{yang2023real,
    title={Real-time Photorealistic Dynamic Scene Representation and Rendering with 4D Gaussian Splatting},
    author={Yang, Zeyu and Yang, Hongye and Pan, Zijie and Zhang, Li},
    booktitle=ICLR,
    year={2023}
}

@inproceedings{wu20234dgs,
    title={4D Gaussian Splatting for Real-Time Dynamic Scene Rendering},
    author={Wu, Guanjun and Yi, Taoran and Fang, Jiemin and Xie, Lingxi and Zhang, Xiaopeng and Wei Wei and Liu, Wenyu and Tian, Qi and Wang Xinggang},
    booktitle=CVPR,
    year={2024}
}

@inproceedings{yang2023deformable,
    title={Deformable 3d gaussians for high-fidelity monocular dynamic scene reconstruction},
    author={Yang, Ziyi and Gao, Xinyu and Zhou, Wen and Jiao, Shaohui and Zhang, Yuqing and Jin, Xiaogang},
    booktitle=CVPR,
    year={2024}
}

@inproceedings{huang2023sc,
    title={{SC-GS}: Sparse-Controlled Gaussian Splatting for Editable Dynamic Scenes},
    author={Huang, Yi-Hua and Sun, Yang-Tian and Yang, Ziyi and Lyu, Xiaoyang and Cao, Yan-Pei and Qi, Xiaojuan},
    booktitle=CVPR,
    year={2024}
}

@inproceedings{lu20243d,
    title={3d geometry-aware deformable gaussian splatting for dynamic view synthesis},
    author={Lu, Zhicheng and Guo, Xiang and Hui, Le and Chen, Tianrui and Yang, Min and Tang, Xiao and Zhu, Feng and Dai, Yuchao},
    booktitle=CVPR, 
    year={2024}
}

@inproceedings{li2023spacetime,
    title={Spacetime Gaussian Feature Splatting for Real-Time Dynamic View Synthesis},
    author={Li, Zhan and Chen, Zhang and Li, Zhong and Xu, Yi},
    booktitle=CVPR, 
    year={2024}
}

@inproceedings{lin2023gaussian,
    title={Gaussian-flow: 4d reconstruction with dynamic 3d gaussian particle},
    author={Lin, Youtian and Dai, Zuozhuo and Zhu, Siyu and Yao, Yao},
    booktitle=CVPR, 
    year={2024}
}

@inproceedings{zwicker2001ewa,
    title={{EWA} volume splatting},
    author={Zwicker, Matthias and Pfister, Hanspeter and Van Baar, Jeroen and Gross, Markus},
    booktitle=VIS,
    year={2001}
}

@inproceedings{xu20234k4d,
    title={4K4D: Real-Time 4D View Synthesis at 4K Resolution},
    author={Xu, Zhen and Peng, Sida and Lin, Haotong and He, Guangzhao and Sun, Jiaming and Shen, Yujun and Bao, Hujun and Zhou, Xiaowei},
    booktitle=CVPR,
    year={2024}
}

@inproceedings{hyperreel,
    title={{HyperReel}: High-fidelity 6-DoF video with ray-conditioned sampling},
    author={Attal, Benjamin and Huang, Jia-Bin and Richardt, Christian and Zollhoefer, Michael and Kopf, Johannes and O’Toole, Matthew and Kim, Changil},
    booktitle=CVPR,
    year={2023}
}

@inproceedings{Sabater2017,
    title={Dataset and Pipeline for Multi-View Light-Field Video},
    author={Sabater, Neus and Boisson, Guillaume and Vandame, Benoit and Kerbiriou, Paul and Babon, Frederic and Hog, Matthieu and Langlois, Tristan and Gendrot, Remy and Bureller, Olivier and Schubert, Arno and Allie, Valerie},
    booktitle=CVPRW,
    year={2017}
}

@article{lombardi2019neural,
  title={Neural volumes: learning dynamic renderable volumes from images},
  author={Lombardi, Stephen and Simon, Tomas and Saragih, Jason and Schwartz, Gabriel and Lehrmann, Andreas and Sheikh, Yaser},
  journal=TOG,
  year={2019}
}

@inproceedings{wang2023mixed,
  title={Mixed neural voxels for fast multi-view video synthesis},
  author={Wang, Feng and Tan, Sinan and Li, Xinghang and Tian, Zeyue and Song, Yafei and Liu, Huaping},
  booktitle=ICCV,
  year={2023}
}

@inproceedings{schonberger2016structure,
  title={Structure-from-motion revisited},
  author={Schonberger, Johannes L and Frahm, Jan-Michael},
  booktitle=CVPR,
  year={2016}
}

@InProceedings{sun20243dgstream,
    author    = {Sun, Jiakai and Jiao, Han and Li, Guangyuan and Zhang, Zhanjie and Zhao, Lei and Xing, Wei},
    title     = {{3DGStream}: On-the-Fly Training of 3D Gaussians for Efficient Streaming of Photo-Realistic Free-Viewpoint Videos},
    booktitle = CVPR,
    year      = {2024}
}

@inproceedings{lee2024fully,
  title={Fully Explict Dynamic Gaussian Splatting},
  author={Lee, Junoh and Jung, Hyunjun and Bae, Inhwan and Won, Changyeon and Jeon, Hae-Gon},
  booktitle=NeurIPS,
  year={2024}
}

@inproceedings{wang2025som,
  title     = {{Shape of Motion}: 4D Reconstruction from a Single Video},
  author    = {Wang, Qianqian and Ye, Vickie and Gao, Hang and Zeng, Weijia and Austin, Jake and Li, Zhengqi and Kanazawa, Angjoo},
  booktitle   = ICCV,
  year      = {2025}
}

@inproceedings{pumarola2020dnerf,
  title={{D-NeRF: Neural Radiance Fields for Dynamic Scenes}},
  author={Pumarola, Albert and Corona, Enric and Pons-Moll, Gerard and Moreno-Noguer, Francesc},
  booktitle=CVPR,
  year={2021}
}

@inproceedings{zhu2024motiongs,
  title={{MotionGS}: Exploring explicit motion guidance for deformable 3d gaussian splatting},
  author={Zhu, Ruijie and Liang, Yanzhe and Chang, Hanzhi and Deng, Jiacheng and Lu, Jiahao and Yang, Wenfei and Zhang, Tianzhu and Zhang, Yongdong},
  booktitle=NeurIPS,
  year={2024}
}

@inproceedings{das2023neural,
	  title={Neural parametric gaussians for monocular non-rigid object reconstruction},
	  author={Das, Devikalyan and Wewer, Christopher and Yunus, Raza and Ilg, Eddy and Lenssen, Jan Eric},
    booktitle=CVPR,
    year={2024}
	}

@InProceedings{kwak2025modecgs,
  title={{MoDec-GS}: Global-to-Local Motion Decomposition and Temporal Interval Adjustment for Compact Dynamic 3D Gaussian Splatting}, 
  author={Sangwoon Kwak and Joonsoo Kim and Jun Young Jeong and Won-Sik Cheong and Jihyong Oh and Munchurl Kim},
  booktitle = CVPR,
  year={2025},
}

@article{huber64,
author = {Peter J. Huber},
title = {{Robust Estimation of a Location Parameter}},
journal = {The Annals of Mathematical Statistics},
year = {1964},
}

@inproceedings{hyung2024effective,
  title={Effective Rank Analysis and Regularization for Enhanced 3D Gaussian Splatting},
  author={HYUNG, JUNHA and Hong, Susung and Hwang, Sungwon and Lee, Jaeseong and Kim, Jin-Hwa and Choo, Jaegul},
  booktitle={Conference on Neural Information Processing Systems (NeurIPS)},
  year={2024},
  organization=NeurIPS
}

@inproceedings{sorkine2007arap,
author = {Sorkine, Olga and Alexa, Marc},
title = {As-rigid-as-possible surface modeling},
year = {2007},
booktitle = SGP,
}

@article{Welford1962Note,
author = {B. P. Welford},
title = {Note on a Method for Calculating Corrected Sums of Squares and Products},
journal = {Technometrics},
volume = {4},
number = {3},
pages = {419--420},
year = {1962}
}

@inproceedings{Xie2025GSLK,
    title={Gaussian Splatting Lucas-Kanade},
    author={Xie, Liuyue and Julin, Joel and Niinuma, Koichiro and Jeni, László A.},
    booktitle=ICLR,
    year={2025}
}

@inproceedings{Li2024ST4DGS,
    author = {Li, Deqi and Huang, Shi-Sheng and Lu, Zhiyuan and Duan, Xinran and Huang, Hua},
    title = {{ST-4DGS}: Spatial-Temporally Consistent 4D Gaussian Splatting for Efficient Dynamic Scene Rendering},
    publisher = {Association for Computing Machinery},
    address = {New York, NY, USA},
    booktitle = {ACM SIGGRAPH 2024 Conference Papers},
    location = {Denver, CO, USA},
    year={2024}
}

@inproceedings{bae2024ed3dgs,
    title={Per-Gaussian Embedding-Based Deformation for Deformable 3D Gaussian Splatting}, 
    author={Bae, Jeongmin and Kim, Seoha and Yun, Youngsik and Lee, Hahyun and Bang, Gun and Uh, Youngjung}, 
    booktitle = ECCV,
    year={2024}
}

@inproceedings{fang2022TiNeuVox,
  author = {Fang, Jiemin and Yi, Taoran and Wang, Xinggang and Xie, Lingxi and Zhang, Xiaopeng and Liu, Wenyu and Nie\ss{}ner, Matthias and Tian, Qi},
  title = {Fast Dynamic Radiance Fields with Time-Aware Neural Voxels},
  year = {2022},
  booktitle = {SIGGRAPH Asia 2022 Conference Papers}
}

@inproceedings{
liu2025modgs,
title={Mo{DGS}: Dynamic Gaussian Splatting from Casually-captured Monocular Videos with Depth Priors},
author={Qingming LIU and Yuan Liu and Jiepeng Wang and Xianqiang Lyu and Peng Wang and Wenping Wang and Junhui Hou},
booktitle=ICLR,
year={2025},
}

@inproceedings{karaev23cotracker,
  title     = {{CoTracker}: It is Better to Track Together},
  author    = {Nikita Karaev and Ignacio Rocco and Benjamin Graham and Natalia Neverova and Andrea Vedaldi and Christian Rupprecht},
  booktitle = ECCV,
  year      = {2024}
}

@inproceedings{Xiao2024SpatialTracker,
    title={{SpatialTracker}: Tracking Any 2D Pixels in 3D Space},
    author={Xiao, Yuxi and Wang, Qianqian and Zhang, Shangzhan and Xue, Nan and Peng, Sida and Shen, Yujun and Zhou, Xiaowei},
    booktitle=CVPR,
    year={2024}
}

@inproceedings{piccinelli2024unidepth,
    title     = {{U}ni{D}epth: Universal Monocular Metric Depth Estimation},
    author    = {Piccinelli, Luigi and Yang, Yung-Hsu and Sakaridis, Christos and Segu, Mattia and Li, Siyuan and Van Gool, Luc and Yu, Fisher},
    booktitle = CVPR,
    year      = {2024}
}

@inproceedings{hu2025DepthCrafter,
            author      = {Hu, Wenbo and Gao, Xiangjun and Li, Xiaoyu and Zhao, Sijie and Cun, Xiaodong and Zhang, Yong and Quan, Long and Shan, Ying},
            title       = {{DepthCrafter}: Generating Consistent Long Depth Sequences for Open-world Videos},
            booktitle   = CVPR,
            year        = {2025}
    }

@inproceedings{teed2020raft,
  title     = {{RAFT}: Recurrent All-Pairs Field Transforms for Optical Flow},
  author    = {Zachary Teed, Jia Deng},
  booktitle = ECCV,
  year      = {2024}
}

@ARTICLE {Ranftl2022midas,
    author  = "Ren\'{e} Ranftl and Katrin Lasinger and David Hafner and Konrad Schindler and Vladlen Koltun",
    title   = "Towards Robust Monocular Depth Estimation: Mixing Datasets for Zero-Shot Cross-Dataset Transfer",
    journal = PAMI,
    year    = "2022",
    volume  = "44",
    number  = "3"
}

@inproceedings{xu2022gmflow,
  title={{GMFlow}: Learning Optical Flow via Global Matching},
  author={Xu, Haofei and Zhang, Jing and Cai, Jianfei and Rezatofighi, Hamid and Tao, Dacheng},
  booktitle=CVPR,
  year={2022}
}

@article{xu2024grid4d,
    title={{Grid4D}: {4D} Decomposed Hash Encoding for High-Fidelity Dynamic Gaussian Splatting},
    author={Jiawei, Xu and Zexin, Fan and Jian, Yang and Jin, Xie},
    journal=NeurIPS,
    year={2024},
}

@inproceedings{hu2025learnable,
  title={Learnable Infinite Taylor Gaussian for Dynamic View Rendering},
  author={Hu, Bingbing and Li, Yanyan and Xie, Rui and Xu, Bo and Dong, Haoye and Yao, Junfeng and Lee, Gim Hee},
  booktitle=CVPR,
  year={2025}
}

@inproceedings{shaw2024swings,
  title={{SWinGS}: sliding windows for dynamic 3d gaussian splatting},
  author={Shaw, Richard and Nazarczuk, Michal and Song, Jifei and Moreau, Arthur and Catley-Chandar, Sibi and Dhamo, Helisa and P{\'e}rez-Pellitero, Eduardo},
  booktitle=ECCV,
  year={2024},
}

@inproceedings{wan2024spgs,
author = {Wan, Diwen and Lu, Ruijie and Zeng, Gang},
title = {Superpoint Gaussian splatting for real-time high-fidelity dynamic scene reconstruction},
year = {2024},
booktitle = ICML,
}

@inproceedings{duan20244drotor,
  title={4d-rotor gaussian splatting: towards efficient novel view synthesis for dynamic scenes},
  author={Duan, Yuanxing and Wei, Fangyin and Dai, Qiyu and He, Yuhang and Chen, Wenzheng and Chen, Baoquan},
  booktitle=SIGGRAPH,
  year={2024}
}

@inproceedings{katsumata2024compact,
  title={A compact dynamic 3d gaussian representation for real-time dynamic view synthesis},
  author={Katsumata, Kai and Vo, Duc Minh and Nakayama, Hideki},
  booktitle=ECCV,
  year={2024},
}

@article{gao2024hicom,
  title={{HiCoM}: Hierarchical coherent motion for dynamic streamable scenes with 3d gaussian splatting},
  author={Gao, Qiankun and Meng, Jiarui and Wen, Chengxiang and Chen, Jie and Zhang, Jian},
  journal=NeurIPS,
  year={2024}
}

@inproceedings{yoon2025splinegs,
  title={{SplineGS}: Learning smooth trajectories in gaussian splatting for dynamic scene reconstruction},
  author={Yoon, Jihwan and Han, Sangbeom and Oh, Jaeseok and Lee, Minsik},
  booktitle=ICLR,
  year={2025}
}

@inproceedings{wuswift4d,
  title={{Swift4D}: Adaptive divide-and-conquer Gaussian Splatting for compact and efficient reconstruction of dynamic scene},
  author={Wu, Jiahao and Peng, Rui and Wang, Zhiyan and Xiao, Lu and Tang, Luyang and Yan, Jinbo and Xiong, Kaiqiang and Wang, Ronggang},
  booktitle=ICLR,
  year={2025}  
}

@inproceedings{liang2025himor,
  title={Himor: Monocular deformable gaussian reconstruction with hierarchical motion representation},
  author={Liang, Yiming and Xu, Tianhan and Kikuchi, Yuta},
  booktitle=CVPR,
  year={2025}
}

@inproceedings{park2025splinegs,
  title={{SplineGS}: Robust motion-adaptive spline for real-time dynamic 3d gaussians from monocular video},
  author={Park, Jongmin and Bui, Minh-Quan Viet and Bello, Juan Luis Gonzalez and Moon, Jaeho and Oh, Jihyong and Kim, Munchurl},
  booktitle=CVPR,
  year={2025}
}

@inproceedings{wang2025freetimegs,
  title={{FreeTimeGS}: Free gaussian primitives at anytime anywhere for dynamic scene reconstruction},
  author={Wang, Yifan and Yang, Peishan and Xu, Zhen and Sun, Jiaming and Zhang, Zhanhua and Chen, Yong and Bao, Hujun and Peng, Sida and Zhou, Xiaowei},
  booktitle=CVPR,
  year={2025}
}

@inproceedings{wu20254dfly,
  title={{4D-Fly}: Fast 4d reconstruction from a single monocular video},
  author={Wu, Diankun and Liu, Fangfu and Hung, Yi-Hsin and Qian, Yue and Zhan, Xiaohang and Duan, Yueqi},
  booktitle=CVPR,
  year={2025}
}

@inproceedings{zhang2025mega,
  title={{Mega}: Memory-efficient 4d gaussian splatting for dynamic scenes},
  author={Zhang, Xinjie and Liu, Zhening and Zhang, Yifan and Ge, Xingtong and He, Dailan and Xu, Tongda and Wang, Yan and Lin, Zehong and Yan, Shuicheng and Zhang, Jun},
  booktitle=ICCV,
  year={2025}
}

@article{lee2026space,
  title={Space-Time Forecasting of Dynamic Scenes with Motion-aware Gaussian Grouping},
  author={Lee, Junmyeong and Choi, Hoseung and Cho, Minsu},
  journal=CVPR,
  year={2026}
}

@inproceedings{yan2023nerfds,
  title={NeRF-DS: Neural Radiance Fields for Dynamic Specular Objects},
  author={Yan, Zhiwen and Li, Chen and Lee, Gim Hee},
  booktitle=CVPR,
  year={2023}
}

@inproceedings{zhang2018perceptual,
  title={The Unreasonable Effectiveness of Deep Features as a Perceptual Metric},
  author={Zhang, Richard and Isola, Phillip and Efros, Alexei A and Shechtman, Eli and Wang, Oliver},
  booktitle={CVPR},
  year={2018}
}

@article{mueller2022instant,
    author = {Thomas M\"uller and Alex Evans and Christoph Schied and Alexander Keller},
    title = {Instant Neural Graphics Primitives with a Multiresolution Hash Encoding},
    journal = {ACM Trans. Graph.},
    volume = {41},
    number = {4},
    year = {2022},
}

@inproceedings{kratimenos2024dynmf,
  title={{DynMF}: Neural motion factorization for real-time dynamic view synthesis with 3d gaussian splatting},
  author={Kratimenos, Agelos and Lei, Jiahui and Daniilidis, Kostas},
  booktitle=ECCV,
  year={2024}
}

@inproceedings{chen2025freegaussian,
  title={FreeGaussian: Annotation-free Control of Articulated Objects via 3D Gaussian Splats with Flow Derivatives}, 
  author={Chen, Qizhi and Qu, Delin and Liu, Junli and Tang, Yiwen and Song, Haoming and Wang, Dong and Zhao, Bin and Li, Xuelong},
  booktitle=AAAI,
  year={2025}
}

@inproceedings{Lu2024scaffoldgs,
  title={Scaffold-gs: Structured 3d gaussians for view-adaptive rendering},
  author={Lu, Tao and Yu, Mulin and Xu, Linning and Xiangli, Yuanbo and Wang, Limin and Lin, Dahua and Dai, Bo},
  booktitle=CVPR,
  year={2024}
}

@article{qi2017pointnetplusplus,
  title={PointNet++: Deep Hierarchical Feature Learning on Point Sets in a Metric Space},
  author={Qi, Charles R and Yi, Li and Su, Hao and Guibas, Leonidas J},
  journal=NeurIPS,
  year={2017}
}

@inproceedings{park2023tidnerf,
  author={Park, Sungheon and Son, Minjung and Jang, Seokhwan and Ahn, Young Chun and Kim, Ji-Yeon and Kang, Nahyup},
  title={Temporal Interpolation is all You Need for Dynamic Neural Radiance Fields}, 
  booktitle=CVPR, 
  year={2023},
}

@article{shin2025chroma,
  title={CHROMA: Consistent Harmonization of Multi-View Appearance via Bilateral Grid Prediction},
  author={Shin, Jisu and Shaw, Richard and Shin, Seunghyun and Zhang, Zhensong and Jeon, Hae-Gon and Perez-Pellitero, Eduardo},
  journal={arXiv preprint arXiv:2507.15748},
  year={2025}
}
